\documentclass[10pt,twocolumn,letterpaper]{article}

\usepackage{iccv}
\usepackage{times}
\usepackage{epsfig}
\usepackage{graphicx}
\usepackage{amsmath}
\usepackage{amssymb}

\usepackage{listings}
\usepackage{color}
\usepackage{subcaption}
\usepackage{mathtools}
\usepackage{paralist} 
\usepackage{colortbl}
\usepackage{nicefrac}
\usepackage{multirow}
\usepackage[table,dvipsnames]{xcolor}
\usepackage{makecell}
\usepackage[outline]{contour}
\usepackage{soul}
\usepackage{bbm}
\usepackage[accsupp]{axessibility}  

\usepackage[percent]{overpic}
\usepackage{xcolor}

\usepackage{appendix}

\usepackage[breaklinks=true,bookmarks=false]{hyperref}

\iccvfinalcopy 

\def\iccvPaperID{4397} 

\ificcvfinal\fi

\begin{document}

\definecolor{yellow}{rgb}{1,1, 0.6}
\definecolor{lightyellow}{rgb}{1,1, 0.8}
\definecolor{orange}{rgb}{1, 0.8, 0.6}
\definecolor{red}{rgb}{1, 0.6, 0.6}

\definecolor{wincolor}{rgb}{0.85, 0.0, 0.0}

\definecolor{darkyellow}{rgb}{0.8, 0.8, 0.5}
\definecolor{darkred}{rgb}{0.7, 0.3, 0.3}
\definecolor{darkgreen}{rgb}{0.3, 0.7, 0.3}
\definecolor{blue}{rgb}{0, 0, 1.0}
\definecolor{green}{rgb}{0, 1.0, 0}
\definecolor{pink}{rgb}{1, 0.4, 0.7}

\newcommand{\barron}[1]{{\color{blue} barron: #1}}
\newcommand{\ricardo}[1]{{\color{darkgreen} ricardo: #1}}
\newcommand{\todo}[1]{{\color{pink} TODO: #1}}

\newcommand{\mbf}[1]{{\mathbf{#1}}}

\let\originalleft\left
\let\originalright\right
\renewcommand{\left}{\mathopen{}\mathclose\bgroup\originalleft}
\renewcommand{\right}{\aftergroup\egroup\originalright}

\newcommand{\norm}[1]{\left\lVert#1\right\rVert}

\newcommand{\expo}[1]{\exp\left(#1\right)}

\newcommand{\modeltheta}{\mathrm{\Theta}}
\newcommand{\absrp}{\sigma}

\newcommand{\numsamples}{N}
\newcommand{\numsamplescoarse}{N_c}
\newcommand{\numsamplesfine}{N_f}
\newcommand{\timenear}{t_n}
\newcommand{\timefar}{t_f}
\newcommand{\deltatime}{\delta}

\newcommand{\posxy}{xy}
\newcommand{\posxyz}{xyz}
\newcommand{\angletheta}{\theta}
\newcommand{\anglephi}{\phi}
\newcommand{\posall}{\posxyz\angletheta\anglephi}

\newcommand{\numfrequencies}{L}

\newcommand{\Ltrain}{\mathcal{L}}
\newcommand{\raybatch}{\mathcal{R}}
\newcommand{\Ccoarse}{\hat{C}_c(\ray)}
\newcommand{\Cfine}{\hat{C}_f(\ray)}
\newcommand{\Ctrue}{C(\ray)}
\newcommand{\pweight}{w}
\newcommand{\normpweight}{\hat{w}}

\newcommand{\scenename}[1]{\textit{#1}}

\newcommand{\shortpara}[1]{{\bf #1}}

\newcommand{\longmodelname}{\emph{multum in parvo} NeRF\xspace}
\newcommand{\shortmodelname}{Mip-NeRF\xspace}
\newcommand{\shortmodelnamelower}{mip-NeRF\xspace}

%
%

\newcommand{\featurename}{integrated positional encoding\xspace}
\newcommand{\shortfeaturename}{IPE\xspace}

\newcommand{\posenc}{positional encoding\xspace}
\newcommand{\shortposenc}{PE\xspace}

\newcommand{\numhidden}{n}

\newcommand{\posencfun}{\gamma}
\newcommand{\coordenc}{\gamma(\mathbf{x})}

\newcommand{\rayorigin}{\mathbf{o}}
\newcommand{\raydir}{\mathbf{d}}
\newcommand{\transpose}{{\operatorname{T}}}

\newcommand\submissionjunk[1]{ #1 }

\newcommand\estimate[1]{\hat{#1}}
\newcommand{\ray}{\mathbf{r}}
\newcommand{\normsq}[1]{\left\lVert#1\right\rVert^2_2}
\newcommand{\position}{\mathbf{x}}
\newcommand{\col}{\mathbf{c}}
\newcommand{\Col}{\mathbf{C}}
\newcommand{\trueCol}{\mathbf{C}^*}
\newcommand{\mlp}{\operatorname{MLP}}
\newcommand{\modelweights}{\Theta}
\newcommand\decay[1]{\alpha \left( #1 \right)}
\newcommand{\basis}{\mathbf{P}}

\newcommand{\density}{\tau}
\newcommand{\baseradius}{\dot r}
\newcommand{\baseangle}{\theta}
\newcommand{\conefun}{\operatorname{F}}
\newcommand{\zval}{t}
\newcommand{\zvec}{\mathbf{\zval}}

\newcommand{\textpyr}[5]{
	\begin{overpic}[width=1.3in]{figures/multiblender_pyrs/#1}
	\put (67,2) {\tiny \sethlcolor{white}\hl{$#2$}}
	\put (84,29) {\tiny \sethlcolor{white}\hl{$#3$}}
	\put (84,18) {\tiny \sethlcolor{white}\hl{$#4$}}
	\put (76,10) {\tiny \sethlcolor{white}\hl{$#5$}}
    \end{overpic}
}

\newcommand{\flatim}[1]{
    \includegraphics[width=0.9in]{figures/multiblender_pyrs/#1}
}

\newcommand{\myparagraph}[1]{\noindent {\bf #1}\,\,\,}

\newcommand{\zc}{\zval_\mu}
\newcommand{\zd}{\zval_\delta}

\newcommand{\dirichlet}{\alpha}
\newcommand{\lossmult}{\lambda}

\newcommand{\numsteps}{n}
\newcommand{\learnrate}{\eta}

\newcommand{\thousand}{K}
\newcommand{\million}{M}

\newcommand{\covmat}{\boldsymbol{\Sigma}}

\title{\!\!\shortmodelname: A Multiscale Representation for Anti-Aliasing Neural Radiance Fields\!}

\author{
Jonathan T. Barron$^1$
\quad
Ben Mildenhall$^1$
\quad
Matthew Tancik$^2$
\\
\quad Peter Hedman$^1$
\quad
Ricardo Martin-Brualla$^1$
\quad
Pratul P. Srinivasan$^1$ \\
\vspace{2mm}
{$^1$Google \quad $^2$UC Berkeley}
}
%

\maketitle
\ificcvfinal\thispagestyle{empty}\fi

\begin{abstract}
The rendering procedure used by neural radiance fields (NeRF) samples a scene with a single ray per pixel and may therefore produce renderings that are excessively blurred or aliased when training or testing images observe scene content at different resolutions. The straightforward solution of supersampling by rendering with multiple rays per pixel is impractical for NeRF, because rendering each ray requires querying a multilayer perceptron hundreds of times. Our solution, which we call ``mip-NeRF'' (à la ``mipmap''), extends NeRF to represent the scene at a continuously-valued scale.
By efficiently rendering anti-aliased conical frustums instead of rays, \shortmodelnamelower reduces objectionable aliasing artifacts and significantly improves NeRF's ability to represent fine details, while also being $7\%$ faster than NeRF and half the size.
Compared to NeRF, \shortmodelnamelower reduces average error rates by $17\%$ on the dataset presented with NeRF and by $60\%$ on a challenging multiscale variant of that dataset that we present. \shortmodelname is also able to match the accuracy of a brute-force supersampled NeRF on our multiscale dataset while being $22\!\times$ faster.
\end{abstract}

\section{Introduction}

Neural volumetric representations such as neural radiance fields (NeRF)~\cite{mildenhall2020} have emerged as a compelling strategy for learning to represent 3D objects and scenes from images for the purpose of rendering photorealistic novel views. Although NeRF and its variants have demonstrated impressive results across a range of view synthesis tasks, NeRF's rendering model is flawed in a manner that can cause excessive blurring and aliasing. NeRF replaces traditional discrete sampled geometry with a continuous volumetric function, parameterized as a multilayer perceptron (MLP) that maps from an input 5D coordinate (3D position and 2D viewing direction) to properties of the scene (volume density and view-dependent emitted radiance) at that location. To render a pixel's color, NeRF casts a single ray through that pixel and out into its volumetric representation, queries the MLP for scene properties at samples along that ray, and composites these values into a single color.

While this approach works well when all training and testing images observe scene content from a roughly constant distance (as done in NeRF and most follow-ups), NeRF renderings exhibit significant artifacts in less contrived scenarios. When the training images observe scene content at multiple resolutions, renderings from the recovered NeRF appear excessively blurred in close-up views and contain aliasing artifacts in distant views. A straightforward solution is to adopt the strategy used in offline raytracing: supersampling each pixel by marching multiple rays through its footprint. But this is prohibitively expensive for neural volumetric representations such as NeRF, which require hundreds of MLP evaluations to render a single ray and several hours to reconstruct a single scene.

\begin{figure}[t]
    \centering
    \includegraphics[width=\linewidth]{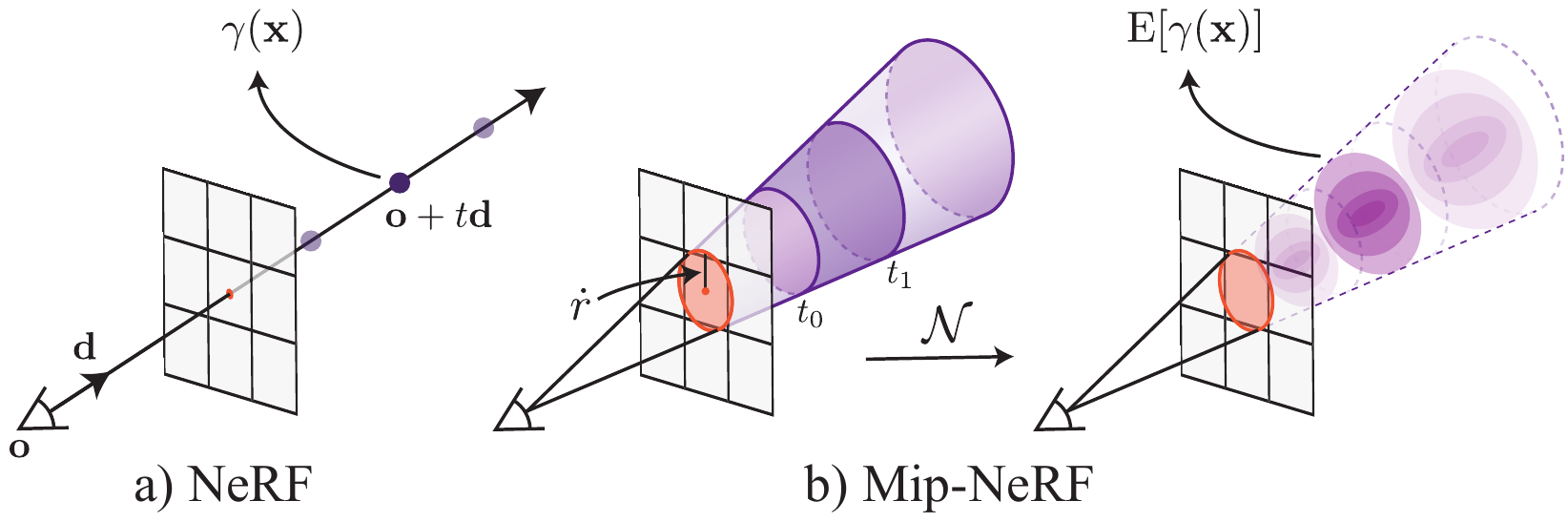}
    \vspace{-0.2in}
    \caption{
    NeRF (a) samples points $\position$ along rays that are traced from the camera center of projection through each pixel, then encodes those points with a positional encoding (\shortposenc) $\posencfun$ to produce a feature $\coordenc$. \shortmodelname (b) instead reasons about the 3D \emph{conical frustum} defined by a camera pixel.
    These conical frustums are then featurized with our \featurename (\shortfeaturename), which works by approximating the frustum with a multivariate Gaussian and then computing the (closed form) integral $\operatorname{E}[\coordenc]$ over the positional encodings of the coordinates within the Gaussian.
    }
    \label{fig:rays}
\end{figure}

In this paper, we take inspiration from the mipmapping approach used to prevent aliasing in computer graphics rendering pipelines. A mipmap represents a signal (typically an image or a texture map) at a set of different discrete downsampling scales and selects the appropriate scale to use for a ray based on the projection of the pixel footprint onto the geometry intersected by that ray. This strategy is known as \emph{pre}-filtering, since the computational burden of anti-aliasing is shifted from render time (as in the brute force supersampling solution) to a precomputation phase--- the  mipmap need only be created once for a given texture, regardless of how many times that texture is rendered.

Our solution, which we call \shortmodelnamelower (\longmodelname, as in ``mipmap''), extends NeRF to simultaneously represent the prefiltered radiance field for a \emph{continuous} space of scales. The input to \shortmodelnamelower is a 3D Gaussian that represents the region over which the radiance field should be integrated. As illustrated in Figure~\ref{fig:rays}, we can then render a prefiltered pixel by querying \shortmodelnamelower at intervals along a cone, using Gaussians that approximate the conical frustums corresponding to the pixel. To encode a 3D position and its surrounding Gaussian region, we propose a new feature representation: an \featurename (\shortfeaturename). This is a generalization of NeRF's \posenc (\shortposenc) that allows a \emph{region} of space to be compactly featurized, as opposed to a single point in space.

\shortmodelname substantially improves upon the accuracy of NeRF, and this benefit is even greater in situations where scene content is observed at different resolutions (\ie setups where the camera moves closer and farther from the scene).
On a challenging multiresolution benchmark we present, \shortmodelnamelower is able to reduce error rates relative to NeRF by $60\%$ on average (see Figure~\ref{fig:aliasing_vis} for visualisations).
\shortmodelname's scale-aware structure also allows us to merge the separate ``coarse'' and ``fine'' MLPs used by NeRF for hierarchical sampling~\cite{mildenhall2020} into a single MLP. As a consequence, \shortmodelnamelower is slightly faster than NeRF ($\sim 7\%$), and has half as many parameters.

\newcommand{\aliasimage}[2]{
	\begin{overpic}[width=0.78in]{figures/aliasing2/#1}
	\put (48,2) {\scriptsize #2}
    \end{overpic}
}

\begin{figure}[t]
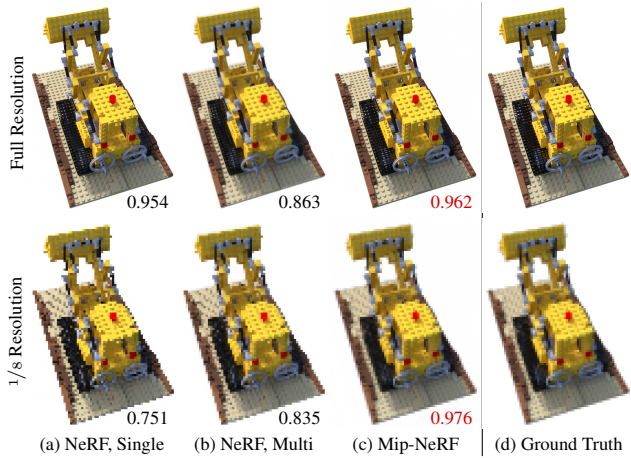

    \centering
    \scriptsize
    \begin{tabular}{@{}c@{}c@{}c@{}c@{}|@{}c@{}}
    \multirow{1}{*}[17ex]{\rotatebox{90}{Full Resolution}} &
    \aliasimage{nerf_full.png}{0.954} & 
    \aliasimage{mnerf_full.png}{0.863} & 
    \aliasimage{mip_full.png}{\color{wincolor} 0.962} & 
    \aliasimage{gt_full.png}{} \\
    \multirow{1}{*}[16.5ex]{\rotatebox{90}{$\nicefrac{1}{8}$ Resolution}} & 
    \aliasimage{nerf_down.png}{0.751} & 
    \aliasimage{mnerf_down.png}{0.835} & 
    \aliasimage{mip_down.png}{\color{wincolor} 0.976} &
    \aliasimage{gt_down.png}{} \\ 
     & (a) NeRF, Single & (b) NeRF, Multi & (c) \shortmodelname & (d) Ground Truth 
    \end{tabular}
    \vspace{-0.1in}
    \caption{
    (a, top) A NeRF trained on full-resolution images is capable of producing photorealistic renderings at new view locations, but \emph{only} at the resolution or scale of the training images. (a, bottom) Pulling the camera back and zooming in (or similarly, adjusting the camera intrinsics to reduce image resolution, as is done here) results in renderings that exhibit severe aliasing. (b) Training a NeRF on multi-resolution images ameliorates this issue slightly but results in poor quality renderings across scales: blur at full resolution, and ``jaggies'' at low resolutions. (c) \shortmodelname, also trained on multi-resolution images, is capable of producing photorealistic renderings across different scales. SSIMs for each image relative to the ground-truth (d) are inset, with the highest SSIM for both scales shown in red.
    }
    \label{fig:aliasing_vis}
\end{figure}

\section{Related Work}

Our work directly extends NeRF~\cite{mildenhall2020}, a highly influential technique for learning a 3D scene representation from observed images in order to synthesize novel photorealistic views.
Here we review the 3D representations used by computer graphics and view synthesis, including recently-introduced continuous neural representations such as NeRF, with a focus on sampling and aliasing.

\myparagraph{Anti-aliasing in Rendering} Sampling and aliasing are fundamental issues that have been extensively studied throughout the development of rendering algorithms in computer graphics. Reducing aliasing artifacts (``anti-aliasing'') is typically done via either supersampling or prefiltering. Supersampling-based techniques~\cite{whitted80} cast multiple rays per pixel while rendering in order to sample closer to the Nyquist frequency. This is an effective strategy to reduce aliasing, but it is expensive, as runtime generally scales linearly with the supersampling rate. Supersampling is therefore typically used only in offline rendering contexts. Instead of sampling more rays to match the Nyquist frequency, prefiltering-based techniques use lowpass-filtered versions of scene content to decrease the Nyquist frequency required to render the scene without aliasing. Prefiltering techniques~\cite{Kaplanyan2016, Neumip, Olano, Lifan2019} are better suited for realtime rendering, because filtered versions of scene content can be precomputed ahead of time, and the correct ``scale'' can be used at render time depending on the target sampling rate. In the context of rendering, prefiltering can be thought of as tracing a cone instead of a ray through each pixel~\cite{amanatides1984ray,igehy1999}: wherever the cone intersects scene content, a precomputed multiscale representation of the scene content (such as a sparse voxel octree~\cite{Heitz2012,laine10} or a mipmap~\cite{williams83}) is queried at the scale corresponding to the cone's footprint. 

Our work takes inspiration from this line of work in graphics and presents a multiscale scene representation for NeRF. Our strategy differs from multiscale representations used in traditional graphics pipelines in two crucial ways. First, we cannot precompute the multiscale representation because the scene's geometry is not known ahead of time in our problem setting --- we are recovering a model of the scene using computer vision, not rendering a predefined CGI asset. 
\shortmodelname therefore must learn a prefiltered representation of the scene  during training.
Second, our notion of scale is continuous instead of discrete. Instead of representing the scene using multiple copies at a fixed number of scales (like in a mipmap), \shortmodelnamelower learns a single neural scene model that can be queried at arbitrary scales.

\myparagraph{Scene Representations for View Synthesis} Various scene representations have been proposed for the task of view synthesis: using observed images of a scene to recover a representation that supports rendering novel photorealistic images of the scene from unobserved camera viewpoints. When images of the scene are captured densely, light field interpolation techniques~\cite{davis12,gortler96,levoy96} can be used to render novel views without reconstructing an intermediate representation of the scene. Issues related to sampling and aliasing have been thoroughly studied within this setting~\cite{chai00}.

Methods that synthesize novel views from sparsely-captured images typically reconstruct explicit representations of the scene's 3D geometry and appearance. Many classic view synthesis algorithms use mesh-based representations along with either diffuse~\cite{waechter2014} or view-dependent~\cite{buehler01, debevec96, wood00} textures. Mesh-based representations can be stored efficiently and are naturally compatible with existing graphics rendering pipelines. However, using gradient-based methods to optimize mesh geometry and topology is typically difficult due to discontinuities and local minima. Volumetric representations have therefore become increasingly popular for view synthesis. Early approaches directly color voxel grids using observed images~\cite{seitz99}, and more recent volumetric approaches use gradient-based learning to train deep networks to predict voxel grid representations of scenes~\cite{flynn19,neuralvolumes,mildenhall2019,sitzmann19,srinivasan19,zhou18}. Discrete voxel-based representations are effective for view synthesis, but they do not scale well to scenes at higher resolutions.

A recent trend within computer vision and graphics research is to replace these discrete representations with \emph{coordinate-based neural representations}, which represent 3D scenes as continuous functions parameterized by MLPs that map from a 3D coordinate to properties of the scene at that location. Some recent methods use coordinate-based neural representations to model scenes as implicit surfaces~\cite{niemeyer2020dvr, yariv2020multiview}, but the majority of recent view synthesis methods are based on the volumetric NeRF representation~\cite{mildenhall2020}. NeRF has inspired many subsequent works that extend its continuous neural volumetric representation for generative modeling~\cite{chan2020pi, schwarz2020graf}, dynamic scenes~\cite{li2021, ost2020neural}, non-rigidly deforming objects~\cite{gafni2020dynamic, park2020deformable}, phototourism settings with changing illumination and occluders~\cite{martinbrualla2020nerfw, tancik2020meta}, and reflectance modeling for relighting~\cite{bi2020, boss2020nerd, nerv2021}.

Relatively little attention has been paid to the issues of sampling and aliasing within the context of view synthesis using coordinate-based neural representations. Discrete representations used for view synthesis, such as polygon meshes and voxel grids, can be efficiently rendered without aliasing using traditional multiscale prefiltering approaches such as mipmaps and octrees. However, coordinate-based neural representations for view synthesis can currently only be anti-aliased using supersampling, which exacerbates their already slow rendering procedure. 
Recent work by Takikawa \etal~\cite{takikawa21} proposes a multiscale representation based on sparse voxel octrees for continuous neural representations of implicit surfaces, but their approach requires that the scene geometry be known a priori, as opposed to our view synthesis setting where the only inputs are observed images. 
\shortmodelname addresses this open problem, enabling the efficient rendering of anti-aliased images during both training and testing as well as the use of multiscale images during training.

\subsection{Preliminaries: NeRF}
\label{sec:prelim_nerf}

NeRF uses the weights of a multilayer perceptron (MLP) to represent a scene as a continuous volumetric field of particles that block and emit light. NeRF renders each pixel of a camera as follows:
A ray $\ray(t) = \rayorigin + t \raydir$ is emitted from the camera's center of projection $\rayorigin$ along the direction $\raydir$ such that it passes through the pixel. A sampling strategy (discussed later) is used to determine a vector of sorted distances $\zvec$ between the camera's predefined near and far planes $t_n$ and $t_f$. For each distance $\zval_k \in \zvec$, we compute its corresponding 3D position along the ray $\position = \ray(t_k)$, then transform each position using a positional encoding:
\small
\begin{equation}
    \gamma(\position)\!=\!\Big[ \sin(\position), \cos(\position), \ldots, \sin\!\big(2^{\numfrequencies-1} \position\big), \cos\!\big(2^{\numfrequencies-1} \position\big) \Big]^\transpose \,. \label{eq:posenc}
\end{equation}
\normalsize
This is simply the concatenation of the sines and cosines of each dimension of the 3D position $\position$ scaled by powers of 2 from $1$ to $2^{\numfrequencies-1}$, where $\numfrequencies$ is a hyperparameter. The fidelity of NeRF depends critically on the use of positional encoding, as it allows the MLP parameterizing the scene to behave as an interpolation function, where $\numfrequencies$ determines the bandwidth of the interpolation kernel (see Tancik \etal \cite{tancik2020fourfeat} for details).
The positional encoding of each ray position $\gamma(\ray(t_k))$ is provided as input to an MLP parameterized by weights $\modelweights$, which outputs a density $\density$ and an RGB color $\col$:
\begin{equation}
    \forall \zval_k \in \zvec, \quad [\density_k, \, \col_k] = \mlp\left( \gamma\left(\ray(t_k)\right); \, \modelweights \right) \, .\label{eq:nerf_mlp}
\end{equation}
The MLP also takes the view direction as input, which is omitted from notation for simplicity.
These estimated densities and colors are used to approximate the volume rendering integral using numerical quadrature, as per Max~\cite{max1995optical}:
\begin{gather}
\Col(\ray; \modelweights, \zvec) = \sum_{k} T_k \left(1-\exp(-\density_k (t_{k+1} - t_k)) \right) \col_k \,, \nonumber \\
    \textrm{with}\quad T_k = \exp \bigg(\!-\! \sum_{k' < k}  \density_{k'} \left(t_{k'+1} - t_{k'} \right)\!\bigg)\,, \label{eq:nerf_render}
\end{gather}
where $\Col(\ray; \modelweights, \zvec)$ is the final predicted color of the pixel.

With this procedure for rendering a NeRF parameterized by $\modelweights$, training a NeRF is straightforward: using a set of observed images with known camera poses, we minimize the sum of squared differences between all input pixel values and all predicted pixel values using gradient descent. To improve sample efficiency, NeRF trains two separate MLPs, one ``coarse'' and one ``fine'', with parameters $\modelweights^c$ and $\modelweights^f$:
\begin{align}
    \underset{\modelweights^c, \modelweights^f}{\operatorname{min}}\, \sum_{\ray \in \mathcal{R}} \Big( & \big\| \trueCol(\ray) - \Col(\ray; \modelweights^c, \zvec^{c}) \big\|_2^2 \label{eq:nerfopt} \\
    + & \big\| \trueCol(\ray) - \Col(\ray; \modelweights^f, \operatorname{sort}(\zvec^{c} \cup \zvec^{f})) \big\|_2^2 \Big)\,, \nonumber
\end{align}
where $\trueCol(\ray)$ is the observed pixel color taken from the input image, and $\mathcal{R}$ is the set of all pixels/rays across all images. Mildenhall \etal construct $\zvec^{c}$ by sampling $64$ evenly-spaced random $\zval$ values with stratified sampling. The compositing weights $w_k = T_k \, \left(1-\exp(-\density_k (t_{k+1} - t_k)) \right)$ produced by the ``coarse'' model are then taken as a piecewise constant PDF describing the distribution of visible scene content, and $128$ new $\zval$ values are drawn from that PDF using inverse transform sampling to produce $\zvec^{f}$. The union of these $192$ $\zval$ values are then sorted and passed to the ``fine'' MLP to produce a final predicted pixel color.

\section{Method}

\begin{figure}[t]
    \centering
    \includegraphics[width=\linewidth]{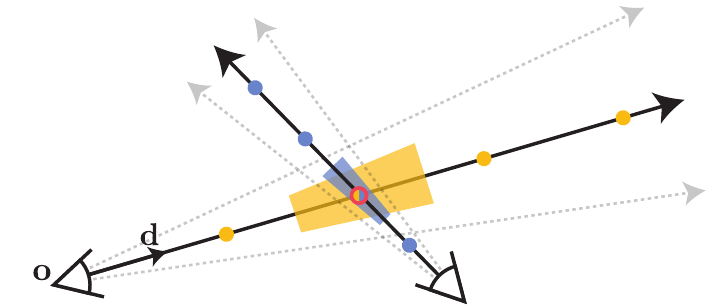}
    \vspace{-0.2in}
    \caption{
    NeRF works by extracting point-sampled positional encoding features (shown here as dots) along each pixel's ray. Those point-sampled features ignore the shape and size of the volume viewed by each ray, so two different cameras imaging the same position at different scales may produce the same ambiguous point-sampled feature, thereby significantly degrading NeRF's performance. In contrast, \shortmodelname casts cones instead of rays and explicitly models the volume of each sampled conical frustum (shown here as trapezoids), thus resolving this ambiguity.
    }
    \label{fig:overlap}
\end{figure}

As discussed, NeRF's point-sampling makes it vulnerable to issues related to sampling and aliasing:
Though a pixel's color is the integration of all incoming radiance within the pixel's frustum, NeRF casts a single infinitesimally narrow ray per pixel, resulting in aliasing. \shortmodelname ameliorates this issue by casting a \emph{cone} from each pixel. Instead of performing point-sampling along each ray, we divide the cone being cast into a series of \emph{conical frustums} (cones cut perpendicular to their axis). And instead of constructing positional encoding (PE) features from an infinitesimally small point in space, we construct an \emph{integrated} positional encoding (IPE) representation of the volume covered by each conical frustum. These changes allow the MLP to reason about the size and shape of each conical frustum, instead of just its centroid. The ambiguity resulting from NeRF's insensitivity to scale and \shortmodelnamelower's solution to this problem are visualized in Figure~\ref{fig:overlap}.
This use of conical frustums and IPE features also allows us to reduce NeRF's two separate ``coarse'' and ``fine'' MLPs into a single multiscale MLP, which increases training and evaluation speed and reduces model size by $50\%$.

\subsection{Cone Tracing and Positional Encoding}

Here we describe \shortmodelnamelower's rendering and featurization procedure, in which we cast a cone and featurize conical frustums along that cone. As in NeRF, images in \shortmodelnamelower are rendered one pixel at a time, so we can describe our procedure in terms of an individual pixel of interest being rendered. For that pixel, we cast a cone from the camera's center of projection $\rayorigin$ along the direction $\raydir$ that passes through the pixel's center.
The apex of that cone lies at $\rayorigin$, and the radius of the cone at the image plane $\rayorigin + \raydir$ is parameterized as $\baseradius$. We set $\baseradius$ to the width of the pixel in world coordinates scaled by $\nicefrac{2}{\sqrt{12}}$, which yields a cone whose section on the image plane has a variance in $x$ and $y$ that matches the variance of the pixel's footprint.
The set of positions $\position$ that lie within a conical frustum between two $\zval$ values $[\zval_0, \zval_1]$ (visualized in Figure~\ref{fig:rays}) is:
\begin{gather}
\conefun(\position, \rayorigin,  \raydir, \baseradius, \zval_0, \zval_1) = 
\mathbbm 1 \Bigg\{ 
      \left(\zval_0 < \frac{\raydir^\transpose (\position - \rayorigin)}{\norm{\raydir}^2_2} < \zval_1 \right) \nonumber \\ 
      \land \, \left(\frac{\raydir^\transpose (\position - \rayorigin)}{\norm{\raydir}_2 \norm{\position - \rayorigin}_2} > \frac{1}{\sqrt{1 + (\baseradius / \norm{\raydir}_2)^2 }}\right) \Bigg\}\,,
\end{gather}
where $\mathbbm 1\{\cdot\}$ is an indicator function: $\conefun(\position, \cdot) = 1$ iff $\position$ is within the conical frustum defined by $(\rayorigin,  \raydir, \baseradius, \zval_0, \zval_1)$.

We must now construct a featurized representation of the volume inside this conical frustum. Ideally, this featurized representation should be of a similar form to the positional encoding features used in NeRF, as Mildenhall~\etal show that this feature representation is critical to NeRF's success~\cite{mildenhall2020}.
There are many viable approaches for this (see the supplement for further discussion) but the simplest and most effective solution we found was to simply compute the expected positional encoding of all coordinates that lie within the conical frustum:
\begin{equation}
    \gamma^*(\rayorigin, \raydir, \baseradius, \zval_0, \zval_1)= \frac{\int \gamma(\position) \conefun(\position, \rayorigin, \raydir, \baseradius, \zval_0, \zval_1)\, d \position}{\int \conefun(\position, \rayorigin, \raydir, \baseradius, \zval_0, \zval_1)\, d \position}\,.
\end{equation}
However, it is unclear how such a feature could be computed efficiently, as the integral in the numerator has no closed form solution. We therefore approximate the conical frustum with a multivariate Gaussian which allows for an efficient approximation to the desired feature, which we will call an ``integrated positional encoding'' (IPE).

To approximate a conical frustum with a multivariate Gaussian, we must compute the mean and covariance of $\conefun(\position, \cdot)$. Because each conical frustum is assumed to be circular, and because conical frustums are symmetric around the axis of the cone, such a Gaussian is fully characterized by three values (in addition to $\rayorigin$ and $\raydir$): the mean distance along the ray $\mu_\zval$, the variance along the ray $\sigma_\zval^2$, and the variance perpendicular to the ray $\sigma_{r}^2$:
\begin{gather}
    \mu_\zval = \zc + \frac{2 \zc \zd^2}{3 \zc^2 + \zd^2}\,,  \quad\,\,
    \sigma_\zval^2 = \frac{\zd^2}{3} -\frac{4 \zd^4 (12 \zc^2 - \zd^2)}{15 (3 \zc^2 + \zd^2)^2}\,, \nonumber \\
    \sigma_{r}^2 = \baseradius^2 \left( \frac{\zc^2}{4} + \frac{5 \zd^2}{12} - \frac{4 \zd^4}{15 
                              (3 \zc^2 + \zd^2)} \right) \,.
\end{gather}
These quantities are parameterized with respect to a midpoint $\zc = (\zval_0 + \zval_1) / 2$ and a half-width $\zd = (\zval_1 - \zval_0) / 2$, which is critical for numerical stability. Please refer to the supplement for a detailed derivation. We can transform this Gaussian from the coordinate frame of the conical frustum into world coordinates as follows:
\begin{equation}
\boldsymbol{\mu} = \rayorigin + \mu_\zval \raydir\,, \quad\,\,
\covmat = \sigma_\zval^2 \left(\raydir\raydir^\transpose\right) + \sigma_{r}^2 \left(\mathbf{I} - \frac{\raydir\raydir^\transpose}{\norm{\raydir}_2^2} \right)\,,
\end{equation}
giving us our final multivariate Gaussian.

Next, we derive the IPE, which is the expectation of a positionally-encoded coordinate distributed according to the aforementioned Gaussian. To accomplish this, it is helpful to first rewrite the PE in Equation~\ref{eq:posenc} as a Fourier feature~\cite{Rahimi2007, tancik2020fourfeat}:
\small 
\begin{equation}
\basis\!=\!\!\begin{bmatrix}
1\!\!&\!\!0\!\!&\!\!0\!\!&\!\!2\!\!&\!\!0\!\!&\!\!0\!\!&\!\!\!\!\!\!&\!\!\!\!2^{L-1}\!\!\!\!&\!\!\!\!0\!\!\!\!&\!\!\!\!0\!\! \\
0\!\!&\!\!1\!\!&\!\!0\!\!&\!\!0\!\!&\!\!2\!\!&\!\!0\!\!&\!\!\!\cdots\!\!\!&\!\!\!\!0\!\!\!\!&\!\!\!\!2^{L-1}\!\!\!\!&\!\!\!\!0\!\!\! \\
0\!\!&\!\!0\!\!&\!\!1\!\!&\!\!0\!\!&\!\!0\!\!&\!\!2\!\!&\!\!\!\!\!\!&\!\!\!0\!\!\!\!&\!\!\!\!0\!\!\!&\!\!\!\!2^{L-1}\!\!\\
\end{bmatrix}^\transpose\!\!\!\!,\,\,\,\gamma(\position)\!=\!\!\begin{bmatrix} \sin(\basis\position) \\ \cos(\basis\position) \end{bmatrix}\,.
\end{equation}
\normalsize
This reparameterization allows us to derive a closed form for IPE.
Using the fact that the covariance of a linear transformation of a variable is a linear transformation of the variable's covariance 
($\operatorname{Cov}[\mathbf{A} \mathbf{x}, \mathbf{B}\mathbf{y}] = \mathbf{A} \operatorname{Cov}[\mathbf{x}, \mathbf{y}] \mathbf{B}^\transpose$) we can identify the mean and covariance of our conical frustum Gaussian after it has been lifted into the PE basis $\basis$:
\begin{equation}
\boldsymbol{\mu}_\gamma = \basis \boldsymbol{\mu}\,, \quad\,\,
\covmat_\gamma = \basis \covmat \basis^\transpose\,.
\end{equation}
The final step in producing an IPE feature is computing the expectation over this lifted multivariate Gaussian, modulated by the sine and the cosine of position. These expectations have simple closed-form expressions:
\begin{align}
    \operatorname{E}_{x \sim \mathcal N(\mu, \sigma^2)}\left[ \sin(x) \right] &= \sin(\mu) \exp\left(-(\nicefrac{1}{2}) \sigma^2 \right)\,, \\
    \operatorname{E}_{x \sim \mathcal N(\mu, \sigma^2)}\left[\cos(x) \right] &= \cos(\mu) \exp\left(-(\nicefrac{1}{2}) \sigma^2 \right) \,.
\end{align}
We see that this expected sine or cosine is simply the sine or cosine of the mean attenuated by a Gaussian function of the variance. With this we can compute our final IPE feature as the expected sines and cosines of the mean and the diagonal of the covariance matrix:
\begin{align}
\gamma(\boldsymbol{\mu}, \covmat) 
&= \operatorname{E}_{\position \sim \mathcal N(\boldsymbol{\mu}_\gamma, \covmat_\gamma)}\left[\gamma(\position)\right] \\
&= \begin{bmatrix} \sin(\boldsymbol{\mu}_\gamma) \circ \exp\left(-(\nicefrac{1}{2}) \operatorname{diag}\left(\covmat_\gamma\right)\right)
\\ \cos(\boldsymbol{\mu}_\gamma) \circ \exp\left(-(\nicefrac{1}{2}) \operatorname{diag}\left(\covmat_\gamma\right)\right)\end{bmatrix}\, , \label{eq:ipe}
\end{align}
where $\circ$ denotes element-wise multiplication. 
Because positional encoding encodes each dimension independently, this expected encoding relies on only the marginal distribution of $\gamma(\position)$, and only the diagonal of the covariance matrix (a vector of per-dimension variances) is needed.
Because $\covmat_\gamma$ is prohibitively expensive to compute due its relatively large size, we directly compute the diagonal of $\covmat_\gamma$:
\begin{equation}
\small \operatorname{diag}(\covmat_\gamma)\!=\!\left[
\operatorname{diag}(\covmat), 4 \operatorname{diag}(\covmat), \ldots, 4^{L-1} \operatorname{diag}(\covmat) \right]^\transpose
\end{equation}
This vector depends on just the diagonal of the 3D position's covariance $\covmat$, which can be computed as:
\begin{equation}
\operatorname{diag}(\covmat) = \sigma_\zval^2 \left(\raydir \circ \raydir\right) + \sigma_{r}^2 \left(\mathbf{1} - \frac{\raydir \circ \raydir}{\norm{\raydir}_2^2} \right)\,.
\end{equation}
If these diagonals are computed directly, IPE features are roughly as expensive as PE features to construct.

\begin{figure}[t]
    \centering
    \includegraphics[width=\linewidth]{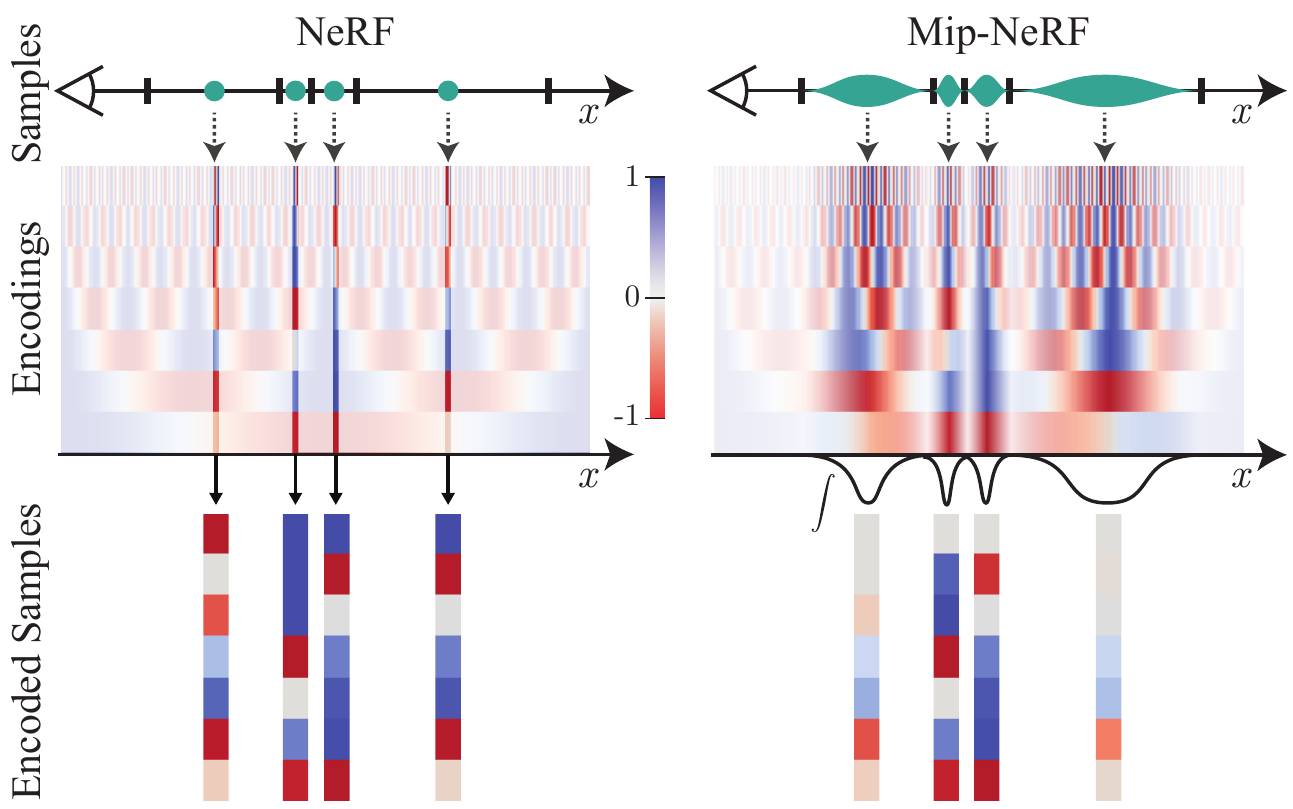}
    \vspace{-0.2in}
    \caption{
    Toy 1D visualizations of the positional encoding (PE) used by NeRF (left) and our integrated positional encoding (IPE) (right). Because NeRF samples points along each ray and encodes all frequencies equally, the high-frequency PE features are aliased, which results in rendering artifacts. By integrating PE features over each interval, the high frequency dimensions of IPE features shrink towards zero when the period of the frequency is small compared to the size of the interval being integrated, resulting in anti-aliased features that implicitly encode the size (and in higher dimensions, the shape) of the interval.
    }
    \label{fig:sample_encoding}
\end{figure}

Figure~\ref{fig:sample_encoding} visualizes the difference between IPE and conventional PE features in a toy 1D domain. IPE features behave intuitively: If a particular frequency in the positional encoding has a period that is larger than the width of the interval being used to construct the IPE feature, then the encoding at that frequency is unaffected. But if the period is smaller than the interval (in which case the PE over that interval will oscillate repeatedly), then the encoding at that frequency is scaled down towards zero. In short, IPE preserves frequencies that are constant over an interval and softly ``removes'' frequencies that vary over an interval, while PE preserves all frequencies up to some manually-tuned hyperparameter $L$. By scaling each sine and cosine in this way, IPE features are effectively \emph{anti-aliased} positional encoding features that smoothly encode the size and shape of a volume of space. IPE also effectively removes $L$ as a hyperparameter: it can simply be set to an extremely large value and then never tuned (see supplement).

\subsection{Architecture}

Aside from cone-tracing and IPE features, \shortmodelnamelower behaves similarly to NeRF, as described in Section~\ref{sec:prelim_nerf}. For each pixel being rendered, instead of a ray as in NeRF, a cone is cast. Instead of sampling $n$ values for $t_k$ along the ray, we sample $n+1$ values for $t_k$, computing IPE features for the interval spanning each adjacent pair of sampled $t_k$ values as previously described. These IPE features are passed as input into an MLP to produce a density $\density_k$ and a color $\col_k$, as in Equation~\ref{eq:nerf_mlp}. Rendering in \shortmodelnamelower follows Equation~\ref{eq:nerf_render}.

Recall that NeRF uses a hierarchical sampling procedure with two distinct MLPs, one ``coarse'' and one ``fine'' (see Equation~\ref{eq:nerfopt}). This was necessary in NeRF because its PE features meant that its MLPs were only able to learn a model of the scene for \emph{one single scale}.
But our cone casting and IPE features allow us to explicitly encode scale into our input features and thereby enable an MLP to learn a \emph{multiscale} representation of the scene. \shortmodelname therefore uses a single MLP with parameters $\modelweights$, which we repeatedly query in a hierarchical sampling strategy. This has multiple benefits: model size is cut in half, renderings are more accurate, sampling is more efficient, and the overall algorithm becomes simpler.
Our optimization problem is:
\small
\begin{equation}
    \underset{\modelweights}{\operatorname{min}}\, \sum_{\ray \in \mathcal{R}} \Big( \lossmult \big\| \trueCol(\ray) - \Col(\ray; \modelweights, \zvec^{c}) \big\|_2^2 + \big\| \trueCol(\ray) - \Col(\ray; \modelweights, \zvec^{f} ) \big\|_2^2 \Big)
\end{equation}
\normalsize
Because we have a single MLP, the ``coarse'' loss must be balanced against the ``fine'' loss, which is accomplished using a hyperparameter $\lossmult$ (we set $\lossmult=0.1$ in all experiments). As in Mildenhall \etal~\cite{mildenhall2020}, our coarse samples $\zvec^{c}$ are produced with stratified sampling, and our fine samples $\zvec^{f}$ are sampled from the resulting alpha compositing weights $\mathbf{w}$ using inverse transform sampling. 
Unlike NeRF, in which the fine MLP is given the sorted union of $64$ coarse samples and $128$ fine samples, in \shortmodelnamelower we simply sample $128$ samples for the coarse model and $128$ samples from the fine model (yielding the same number of total MLP evaluations as in NeRF, for fair comparison). Before sampling $\zvec^{f}$, we modify the weights $\mathbf{w}$ slightly:
\begin{equation}
w'_k = \frac{1}{2}\left(\max(w_{k-1}, w_{k}) + \max(w_{k}, w_{k+1})\right) + \dirichlet\,.
\end{equation}
We filter $\mathbf{w}$ with a 2-tap max filter followed by a 2-tap blur filter (a ``blurpool''~\cite{zhang2019making}), which produces a wide and smooth upper envelope on $\mathbf{w}$. A hyperparameter $\dirichlet$ is added to that envelope before it is re-normalized to sum to $1$, which ensures that some samples are drawn even in empty regions of space (we set $\dirichlet=0.01$ in all experiments).

\shortmodelname is implemented on top of JaxNeRF~\cite{jaxnerf2020github}, a JAX~\cite{jax2018github} reimplementation of NeRF that achieves better accuracy and trains faster than the original TensorFlow implementation.
We follow NeRF's training procedure: 1 million iterations of Adam~\cite{adam} with a batch size of $4096$ and a learning rate that is annealed logarithmically from $5 \cdot 10^{-4}$ to $5 \cdot 10^{-6}$.
See the supplement for additional details and some additional differences between JaxNeRF and \shortmodelnamelower that do not affect performance significantly and are incidental to our primary contributions: cone-tracing, IPE, and the use of a single multiscale MLP.

\section{Results}

\begin{figure*}[t!]
    \centering
    \begin{tabular}{@{}c@{\,\,}c|c|c|c@{}}
    \flatim{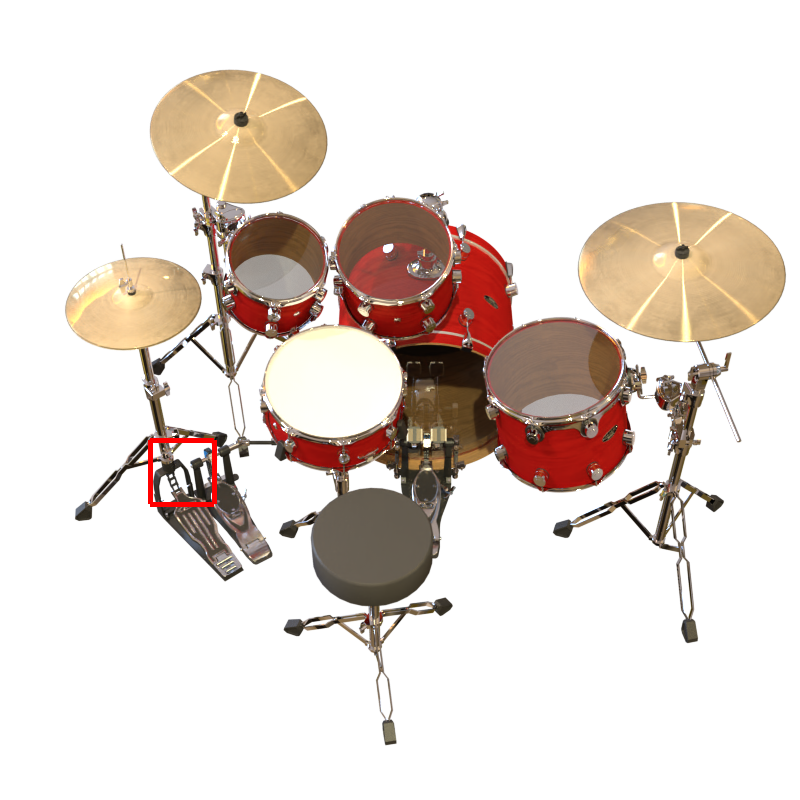} & 
    \textpyr{gt_drums_0_440_150_64.png}{}{}{}{} & 
    \textpyr{jaxnerf_multiblender_drums_0_440_150_64.png}{0.709}{0.910}{0.931}{0.663} & 
    \textpyr{jaxnerf_multiblender_extras_drums_0_440_150_64.png}{0.863}{0.959}{0.971}{0.881} & 
    \textpyr{prenerf_multiblender_drums_0_440_150_64.png}{ \color{wincolor} 0.940}{ \color{wincolor} 0.979}{ \color{wincolor} 0.989}{ \color{wincolor} 0.978} \\ 
    \flatim{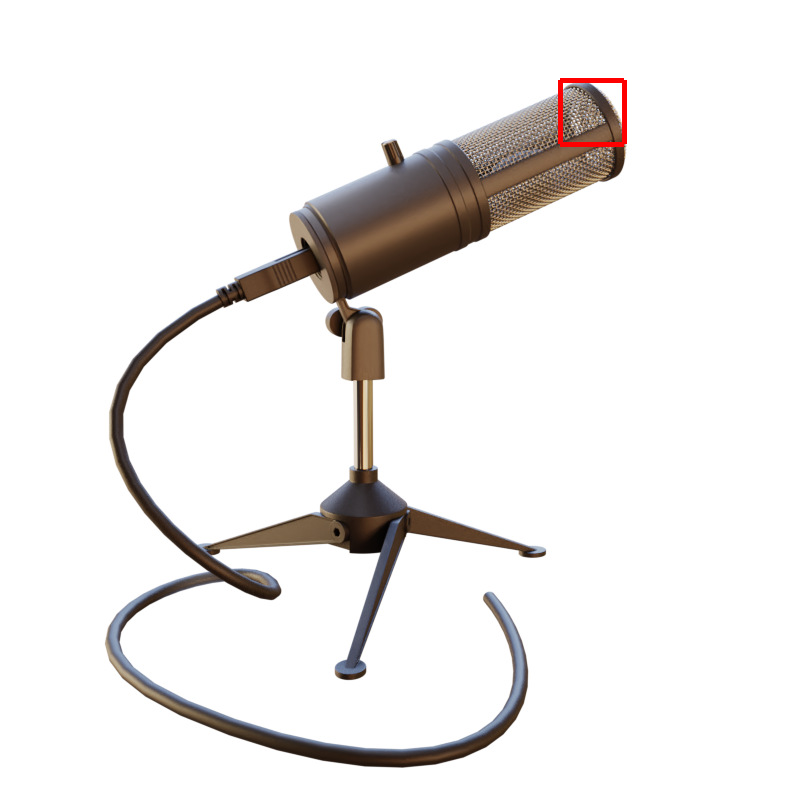} & 
    \textpyr{gt_mic_120_80_560_64.png}{}{}{}{} & 
    \textpyr{jaxnerf_multiblender_mic_120_80_560_64.png}{0.448}{0.562}{0.696}{0.906} & 
    \textpyr{jaxnerf_multiblender_extras_mic_120_80_560_64.png}{0.525}{0.633}{0.794}{0.918} & 
    \textpyr{prenerf_multiblender_mic_120_80_560_64.png}{ \color{wincolor} 0.785}{ \color{wincolor} 0.837}{ \color{wincolor} 0.861}{ \color{wincolor} 0.975} \\ 
    \multicolumn{2}{c|}{\footnotesize{Ground-Truth}} &
    \footnotesize{NeRF} & 
    \footnotesize{NeRF + Area, Center, Misc} & 
    \footnotesize{\shortmodelname}
    \end{tabular}
    \vspace{-0.1in}
    \caption{
    Visualizations of the output of \shortmodelnamelower compared to the ground truth, NeRF, and an improved version of NeRF on test set images from two scenes in our multiscale Blender dataset.
    We visualize a cropped region of both scenes at 4 different scales, displayed as an image pyramid with the SSIM for each scale shown to its lower right and with the highest SSIM at each scale highlighted in red. \shortmodelname outperforms NeRF and its improved version by a significant margin, both visually and quantitatively. See the supplement for more such visualizations.
    }
    \label{fig:multi_vis}
\end{figure*}

\begin{table*}[]
    \centering
    \resizebox{\linewidth}{!}{
    \begin{tabular}{@{}l@{\,\,}|cccc|cccc|cccc|c|c@{\,\,}c@{}}
    & \multicolumn{4}{c|}{PSNR $\uparrow$} & \multicolumn{4}{c|}{SSIM $\uparrow$} & \multicolumn{4}{c|}{LPIPS $\downarrow$} & &  & \\
    & Full Res. & $\nicefrac{1}{2}$ Res. & $\nicefrac{1}{4}$ Res. & $\nicefrac{1}{8}$ Res. & Full Res. & $\nicefrac{1}{2}$ Res. & $\nicefrac{1}{4}$ Res. & $\nicefrac{1}{8}$ Res. & Full Res. & $\nicefrac{1}{2}$ Res. & $\nicefrac{1}{4}$ Res. & $\nicefrac{1}{8}$ Res & Avg. $\downarrow$ & Time (hours) & \# Params \\ \hline
    NeRF (Jax Impl.)~\cite{jaxnerf2020github,mildenhall2020} &                    31.196 &                    30.647 &                    26.252 &                    22.533 &                    0.9498 &                    0.9560 &                    0.9299 &                    0.8709 &                    0.0546 &                    0.0342 &                    0.0428 &                    0.0750 &                    0.0288 & 3.05 $\pm$ 0.04 & 1,191\thousand \\
NeRF + Area Loss                                                  &                    27.224 &                    29.578 &                    29.445 &                    25.039 &                    0.9113 &                    0.9394 &                    0.9524 &                    0.9176 &                    0.1041 &                    0.0677 &                    0.0406 &                    0.0469 &                    0.0305 & 3.03 $\pm$ 0.03 & 1,191\thousand \\
NeRF + Area, Centered Pixels                                      &                    29.893 &                    32.118 &                    33.399 &                    29.463 &                    0.9376 &                    0.9590 &                    0.9728 &                    0.9620 &                    0.0747 &                    0.0405 &                    0.0245 &                    0.0398 &                    0.0191 & 3.02 $\pm$ 0.05 & 1,191\thousand \\
NeRF + Area, Center, Misc.                                        &                    29.900 &                    32.127 &                    33.404 &                    29.470 &                    0.9378 &                    0.9592 &                    0.9730 &                    0.9622 &                    0.0743 &                    0.0402 &                    0.0243 &                    0.0394 &                    0.0190 & 2.94 $\pm$ 0.02 & 1,191\thousand \\
\hline
\shortmodelname                                                   & \cellcolor{orange} 32.629 & \cellcolor{red}    34.336 & \cellcolor{orange} 35.471 & \cellcolor{yellow} 35.602 & \cellcolor{orange} 0.9579 & \cellcolor{orange} 0.9703 & \cellcolor{orange} 0.9786 & \cellcolor{orange} 0.9833 & \cellcolor{orange} 0.0469 & \cellcolor{yellow} 0.0260 & \cellcolor{orange} 0.0168 & \cellcolor{orange} 0.0120 & \cellcolor{red}    0.0114 & 2.84 $\pm$ 0.01 & 612\thousand \\
\shortmodelname w/o Misc.                                         & \cellcolor{yellow} 32.610 & \cellcolor{orange} 34.333 & \cellcolor{red}    35.497 & \cellcolor{orange} 35.638 & \cellcolor{yellow} 0.9577 & \cellcolor{orange} 0.9703 & \cellcolor{red}    0.9787 & \cellcolor{red}    0.9834 & \cellcolor{yellow} 0.0470 & \cellcolor{orange} 0.0259 & \cellcolor{red}    0.0167 & \cellcolor{orange} 0.0120 & \cellcolor{red}    0.0114 & 2.82 $\pm$ 0.03 & 612\thousand \\
\shortmodelname w/o Single MLP                                    &                    32.401 &                    34.131 & \cellcolor{yellow} 35.462 & \cellcolor{red}    35.967 &                    0.9566 & \cellcolor{yellow} 0.9693 & \cellcolor{yellow} 0.9780 & \cellcolor{red}    0.9834 &                    0.0479 &                    0.0268 & \cellcolor{yellow} 0.0169 & \cellcolor{red}    0.0116 & \cellcolor{orange} 0.0115 & 3.40 $\pm$ 0.01 & 1,191\thousand \\
\shortmodelname w/o Area Loss                                     & \cellcolor{red}    33.059 & \cellcolor{yellow} 34.280 &                    33.866 &                    30.714 & \cellcolor{red}    0.9605 & \cellcolor{red}    0.9704 &                    0.9747 & \cellcolor{yellow} 0.9679 & \cellcolor{red}    0.0427 & \cellcolor{red}    0.0256 &                    0.0213 & \cellcolor{yellow} 0.0308 & \cellcolor{yellow} 0.0139 & 2.82 $\pm$ 0.01 & 612\thousand \\
\shortmodelname w/o \shortfeaturename                             &                    29.876 &                    32.160 &                    33.679 &                    29.647 &                    0.9384 &                    0.9602 &                    0.9742 &                    0.9633 &                    0.0742 &                    0.0393 &                    0.0226 &                    0.0378 &                    0.0186 & 2.79 $\pm$ 0.01 & 612\thousand \\
    \end{tabular}
    }
    \vspace{-0.1in}
    \caption{
    A quantitative comparison of \shortmodelnamelower and its ablations against NeRF and several NeRF variants on the test set of our multiscale Blender dataset. See the text for details. 
    }
    \label{tab:avg_multiblender_results}
\end{table*}

We evaluate \shortmodelnamelower on the Blender dataset presented in the original NeRF paper~\cite{mildenhall2020} and also on a simple multiscale variant of that dataset designed to better probe accuracy on multi-resolution scenes and to highlight NeRF's critical vulnerability on such tasks. 
We report the three error metrics used by NeRF: PSNR, SSIM~\cite{wang2004image}, and LPIPS~\cite{zhang2018unreasonable}. To enable easier comparison, we also present an ``average'' error metric that summarizes all three metrics: the geometric mean of  $\mathrm{MSE}=10^{-\mathrm{PSNR}/10}$, $\sqrt{1-\mathrm{SSIM}}$ (as per~\cite{brunet2011mathematical}), and $\mathrm{LPIPS}$. We additionally report runtimes (median and median absolute deviation of wall time) as well as the number of network parameters for each variant of NeRF and \shortmodelnamelower. All JaxNeRF and \shortmodelnamelower experiments are trained on a TPU v2 with 32 cores~\cite{jouppi2017datacenter}.

We constructed our multiscale Blender benchmark because the original Blender dataset used by NeRF has a subtle but critical weakness: all cameras have the same focal length and resolution and are placed at the same distance from the object. As a result, this Blender task is significantly easier than most real-world datasets, where cameras may be more close or more distant from the subject or may zoom in and out. The limitation of this dataset is complemented by the limitations of NeRF: despite NeRF's tendency to produce aliased renderings,
it is able to produce excellent results on the Blender dataset because that dataset systematically avoids this failure mode.

\myparagraph{Multiscale Blender Dataset} Our multiscale Blender dataset is a straightforward modification to NeRF's Blender dataset, designed to probe aliasing and scale-space reasoning. This dataset was constructed by taking each image in the Blender dataset, box downsampling it a factor of $2$, $4$, and $8$ (and modifying the camera intrinsics accordingly), and combining the original images and the three downsampled images into one single dataset. Due to the nature of projective geometry, this is similar to re-rendering the original dataset where the distance to the camera has been increased by scale factors of $2$, $4$, and $8$. When training \shortmodelnamelower on this dataset, we scale the loss of each pixel by the area of that pixel's footprint in the original image (the loss for pixels from the $\nicefrac{1}{4}$ images is scaled by $16$, etc) so that the few low-resolution pixels have comparable influence to the many high-resolution pixels. The average error metric for this task uses the arithmetic mean of each error metric across all four scales. 

The performance of \shortmodelnamelower for this multiscale dataset can be seen in Table~\ref{tab:avg_multiblender_results}. Because NeRF is the state of the art on the Blender dataset (as will be shown in Table~\ref{tab:avg_blender_results}), we evaluate against only NeRF and several improved versions of NeRF: ``Area Loss'' adds the aforementioned scaling of the loss function by pixel area used by \shortmodelnamelower, ``Centered Pixels'' adds a  half-pixel offset added to each ray's direction such that rays pass through the center of each pixel (as opposed to the corner of each pixel as was done in Mildenhall \etal) and ``Misc'' adds some small changes that slightly improve the stability of training (see supplement).  We also evaluate against several ablations of \shortmodelnamelower: ``w/o Misc'' removes those small changes, ``w/o Single MLP'' uses NeRF's two-MLP training scheme from Equation~\ref{eq:nerfopt}, ``w/o Area Loss'' removes the loss scaling by pixel area, and ``w/o IPE'' uses PE instead of IPE, which causes \shortmodelnamelower to use NeRF's ray-casting (with centered pixels) instead of our cone-casting.

\shortmodelname reduces average error by $60\%$ on this task and outperforms NeRF by a large margin on all metrics and all scales. ``Centering'' pixels improves NeRF's performance substantially, but not enough to approach \shortmodelnamelower. Removing IPE features causes \shortmodelnamelower's performance to degrade to the performance of ``Centered'' NeRF, thereby demonstrating that cone-casting and IPE features are the primary factors driving performance (though the area loss contributes substantially). The ``Single MLP'' \shortmodelnamelower ablation performs well but has twice as many parameters and is nearly $20\%$ slower than \shortmodelnamelower (likely due to this ablation's need to sort $\zval$ values and poor hardware throughput due to its changing tensor sizes across its ``coarse'' and ``fine'' scales). \shortmodelname is also $\sim\!7\%$ faster than NeRF. See Figure~\ref{fig:multi_vis} and the supplement for visualizations.

\newcommand{\textimage}[2]{
	\begin{overpic}[width=0.77in]{figures/blender_tiles/#1}
	\put (71,2) {\tiny \sethlcolor{white}\hl{$#2$}}
    \end{overpic}
}

\begin{figure}[t]
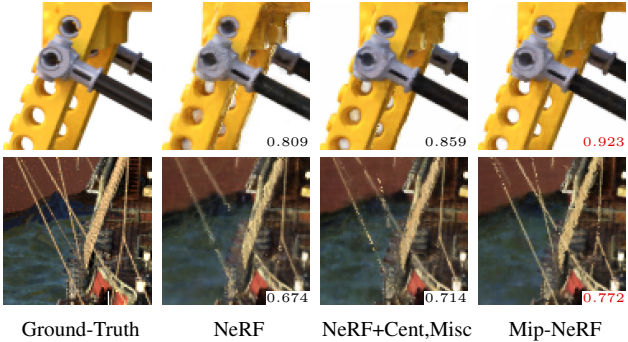

    \centering
    \begin{tabular}{@{}c@{\,}c@{\,}c@{\,}c@{}}
		\textimage{lego_080_gt.png}{} &
        \textimage{lego_080_jaxnerf_blender_image.png}{0.809} &
        \textimage{lego_080_jaxnerf_blender_center_extras_image.png}{0.859} &
        \textimage{lego_080_prenerf_blender_image.png}{\color{wincolor} 0.923} \\
		\textimage{ship_075_gt_alt.png}{} &
        \textimage{ship_075_jaxnerf_blender_image_alt.png}{0.674} &
        \textimage{ship_075_jaxnerf_blender_center_extras_image_alt.png}{0.714} &
        \textimage{ship_075_prenerf_blender_image_alt.png}{\color{wincolor} 0.772} \\
        \footnotesize{Ground-Truth} &
        \footnotesize{NeRF} & 
        \footnotesize{NeRF+Cent,Misc} & 
        \footnotesize{\shortmodelname}
    \end{tabular}
    \vspace{-0.1in}
    \caption{
    Even on the less challenging single-scale Blender dataset of Mildenhall \etal~\cite{mildenhall2020}, \shortmodelnamelower significantly outperforms NeRF and our improved version of NeRF, particularly on small or thin objects such as the holes of the LEGO truck (top) and the ropes of the ship (bottom).
    }
    \label{fig:single_vis}
\end{figure}

\begin{table}[]
    \centering
    \resizebox{\linewidth}{!}{
    \begin{tabular}{@{}l|cccc|cc@{}}
     & \!PSNR $\uparrow$\! & \!SSIM $\uparrow$\! & \!LPIPS $\downarrow$\! & \!Avg. $\downarrow$\! & Time (hours) & \# Params \\ \hline
SRN~\cite{srn}                                                    &                    22.26  &                    0.846  &                    0.170  & 0.0735 &                   -  &  -  \\
Neural Volumes~\cite{neuralvolumes}                               &                    26.05  &                    0.893  &                    0.160  & 0.0507 &               -  &  -  \\
LLFF~\cite{mildenhall2019}                                        &                    24.88  &                    0.911  &                    0.114  & 0.0480 &             \textcolor{gray}{$\sim\!$0.16} &  -  \\
NSVF~\cite{liu2020neural} & 31.74 & 0.953 & 0.047 & 0.0190 & - & 3.2\million\, -  16\million \\
NeRF (TF Impl.)~\cite{mildenhall2020}                    &                    31.01  &                    0.947  &                    0.081  & 0.0245 &               \textcolor{gray}{$>\!$12} & 1,191\thousand \\
    NeRF (Jax Impl.)~\cite{jaxnerf2020github,mildenhall2020} &                    31.74 &                    0.953 &                    0.050 &                    0.0194 & $3.05 \pm 0.01$ & 1,191\thousand \\
NeRF + Centered Pixels                                            &                    32.30 &                    0.957 &                    0.046 &                    0.0178 & $2.99 \pm 0.06$ & 1,191\thousand \\
NeRF + Center, Misc.                                              &                    32.28 &                    0.957 &                    0.046 &                    0.0178 & $3.06 \pm 0.03$ & 1,191\thousand \\
\hline
\shortmodelname                                                   & \cellcolor{red}    33.09 & \cellcolor{red}    0.961 & \cellcolor{red}    0.043 & \cellcolor{red}    0.0161 & $2.89 \pm 0.00$ & 612\thousand \\
\shortmodelname w/o Misc.                                         & \cellcolor{orange} 33.04 & \cellcolor{orange} 0.960 & \cellcolor{red}    0.043 & \cellcolor{red}    0.0162 & $2.89 \pm 0.01$ & 612\thousand \\
\shortmodelname w/o Single MLP                                    & \cellcolor{yellow} 32.71 & \cellcolor{yellow} 0.959 & \cellcolor{orange} 0.044 & \cellcolor{orange} 0.0168 & $3.63 \pm 0.02$ & 1,191\thousand \\
\shortmodelname w/o \shortfeaturename                             &                    32.48 &                    0.958 & \cellcolor{yellow} 0.045 & \cellcolor{yellow} 0.0173 & $2.84 \pm 0.00$ & 612\thousand \\

    \end{tabular}
    }
    \vspace{-0.1in}
    \caption{
    A comparison of \shortmodelnamelower and its ablations against several baseline algorithms and variants of NeRF on the single-scale Blender dataset of Mildenhall~\etal~\cite{mildenhall2020}. Training times taken from prior work (when available) are indicated in gray, as they are not directly comparable.
    }
    \label{tab:avg_blender_results}
\end{table}

\myparagraph{Blender Dataset} Though the sampling issues that \shortmodelnamelower was designed to fix are most prominent in the Multiscale Blender dataset, \shortmodelnamelower also outperforms NeRF on the easier single-scale Blender dataset presented in Mildenhall~\etal~\cite{mildenhall2020}, as shown in Table~\ref{tab:avg_blender_results}. We evaluate against the baselines used by NeRF, NSVF~\cite{liu2020neural}, and the same variants and ablations that were used previously (excluding ``Area Loss'', which is not used by \shortmodelnamelower for this task). Though less striking than the multiscale Blender dataset, \shortmodelnamelower is able to reduce average error by $\sim\!17\%$ compared to NeRF while also being faster. This improvement in performance is most visually apparent in challenging cases such as small or thin structures, as shown in Figure~\ref{fig:single_vis}.

\myparagraph{Supersampling}
As discussed in the introduction, \shortmodelnamelower is a prefiltering approach for anti-aliasing. An alternative approach is supersampling, which can be accomplished in NeRF by casting multiple jittered rays per pixel. Because our multiscale dataset consists of downsampled versions of full-resolution images, we can construct a ``supersampled NeRF'' by training a NeRF (the ``NeRF + Area, Center, Misc.'' variant that performed best previously) using \emph{only} full-resolution images, and then rendering \emph{only} full-resolution images, which we then manually downsample. 
This baseline has an unfair advantage: we manually remove the low-resolution images in the multiscale dataset, which would otherwise degrade NeRF's performance as previously demonstrated. This strategy is not viable in most real-world datasets, as it is usually not possible to known a-priori which images correspond to which scales of image content.
Despite this baseline's advantage, \shortmodelnamelower matches its accuracy while being $\sim\!22\times$ faster (see Table~\ref{tab:supersampling}).

\begin{table}[t]
    \centering
    \resizebox{\linewidth}{!}{
    \begin{tabular}{@{}l|cccc|c|r}
    & \multicolumn{5}{c|}{PSNR $\uparrow$} & \multicolumn{1}{c}{Avg. Time} \\
    & Full Res. & $\nicefrac{1}{2}$ Res. & $\nicefrac{1}{4}$ Res. & $\nicefrac{1}{8}$ Res. & Mean & \multicolumn{1}{c}{(sec./MP)} \\ \hline
     NeRF + Area, Center, Misc.                                             & \cellcolor{yellow} 29.90 &                    32.13 &                    33.40 &                    29.47 &                    31.23 & \cellcolor{orange} 2.61\quad\quad \\
SS NeRF + Area, Center, Misc.                                          & \cellcolor{orange} 32.25 & \cellcolor{yellow} 34.27 & \cellcolor{orange} 35.99 & \cellcolor{orange} 35.73 & \cellcolor{orange} 34.56 &                    55.52\quad\quad \\
\hline
\shortmodelname                                                        & \cellcolor{red}    32.60 & \cellcolor{orange} 34.30 & \cellcolor{yellow} 35.41 & \cellcolor{yellow} 35.55 & \cellcolor{yellow} 34.46 & \cellcolor{red}    2.48\quad\quad \\
SS \shortmodelname                                                     & \cellcolor{red}    32.60 & \cellcolor{red}    34.78 & \cellcolor{red}    36.59 & \cellcolor{red}    36.16 & \cellcolor{red}    35.03 & \cellcolor{yellow} 52.75\quad\quad \\
    \end{tabular}
    }
    \vspace{-0.1in}
    \caption{
    A comparison of \shortmodelnamelower and our improved NeRF variant where both algorithms are supersampled (``SS''). \shortmodelname nearly matches the accuracy of ``SS NeRF'' while being $22\times$ faster. Adding supersampling to \shortmodelnamelower improves its accuracy slightly. We report times for rendering the test set, normalized to seconds-per-megapixel (training times are the same as Tables~\ref{tab:avg_multiblender_results} and \ref{tab:avg_blender_results}).
    }
    \label{tab:supersampling}
\end{table}

\section{Conclusion}

We have presented \shortmodelnamelower, a multiscale NeRF-like model that addresses the inherent aliasing of NeRF. NeRF works by casting rays, encoding the positions of points along those rays, and training separate neural networks at distinct scales. In contrast, \shortmodelnamelower casts \emph{cones}, encodes the positions \emph{and sizes} of conical frustums, and trains a \emph{single} neural network that models the scene at multiple scales. 
By reasoning explicitly about sampling and scale, \shortmodelnamelower is able to reduce error rates relative to NeRF by $60\%$ on our own multiscale dataset, and by $17\%$ on NeRF's single-scale dataset, while also being $7\%$ faster than NeRF.
\shortmodelname is also able to match the accuracy of a brute-force supersampled NeRF variant, while being $22 \times$ faster.
We hope that the general techniques presented here will be valuable to other researchers working to improve the performance of raytracing-based neural rendering models.

\paragraph{Acknowledgements}
We thank Janne Kontkanen and David Salesin for their comments on the text, Paul Debevec for constructive discussions, and Boyang Deng for JaxNeRF. MT is funded by an NSF Graduate Fellowship.

\clearpage

{\small
\bibliographystyle{ieee_fullname}
\bibliography{bib}
}

\clearpage
\appendix

\section{Conical Frustum Integral Derivations}

In order to derive formulas for the various moments of the uniform distribution over a conical frustum, we consider an axis-aligned cone parameterized as $(x,y,z) = \varphi(r, \zval, \theta) = (r \zval \cos\theta, r \zval \sin \theta, \zval)$ for $\theta \in [0, 2\pi)$, $\zval \geq 0$, $|r| \leq \baseradius$. 
This change of variables from Cartesian space gives us a differential term:
\begin{align}
    dx \, dy \, dz &= |\det(D\varphi)(r, \zval, \theta)| dr\, d\zval \, d\theta \\
    &= 
    \begin{vmatrix}
    t \cos \theta & t \sin \theta & 0 \\
    r \cos \theta & r \sin \theta & 1 \\
    -rt\sin\theta & rt\cos\theta & 0 \\
    \end{vmatrix} dr\, d\zval \, d\theta \\
    &= 
    (rt^2 \cos^2 \theta + rt^2 \sin\theta)dr\, d\zval \, d\theta \\
    &= rt^2 dr\, d\zval \, d\theta \, .
\end{align}

The volume of the conical frustum (which serves as the normalizing constant for the uniform distribution) is:
\begin{align}
    V &= \int_{0}^{2\pi} \int_{\zval_0}^{\zval_1} \int_{0}^{\baseradius} r \zval^2 \, dr \, d\zval \, d\theta \\
    &= \frac{\baseradius^2}{2} \cdot \frac{\zval_1^3 - \zval_0^3}{3}\cdot 2\pi \\
    &= \pi \baseradius^2 \frac{\zval_1^3 - \zval_0^3}{3} 
\end{align}
Thus the probability density function for points uniformly sampled from the conical frustum is $rt^2 / V$.
The first moment of $\zval$ is:
\begin{align}
    E[\zval] &= \frac{1}{V} \int_{0}^{2\pi} \int_{\zval_0}^{\zval_1} \int_{0}^{\baseradius} \zval \cdot r \zval^2 \, dr \, d\zval \, d\theta \\
    &= \frac{1}{V} \int_{0}^{2\pi} \int_{\zval_0}^{\zval_1} \int_{0}^{\baseradius}  r \zval^3 \, dr \, d\zval \, d\theta \\
    &= \frac{1}{V} \cdot \pi \baseradius^2 \frac{\zval_1^4 - \zval_0^4}{4} \\
    &= \frac{3(\zval_1^4 - \zval_0^4)}{4(\zval_1^3 - \zval_0^3)} \, .
\end{align}
The moments of $x$ and $y$ are both zero by symmetry.
The second moment of $\zval$ is
\begin{align}
    E[\zval^2] &= \frac{1}{V} \int_{0}^{2\pi} \int_{\zval_0}^{\zval_1} \int_{0}^{\baseradius} \zval^2 \cdot r \zval^2 \, dr \, d\zval \, d\theta \\
    &= \frac{1}{V} \int_{0}^{2\pi} \int_{\zval_0}^{\zval_1} \int_{0}^{\baseradius}  r \zval^4 \, dr \, d\zval \, d\theta \\
    &= \frac{1}{V} \cdot \pi \baseradius^2 \frac{\zval_1^5 - \zval_0^5}{5} \\
    &= \frac{3(\zval_1^5 - \zval_0^5)}{5(\zval_1^3 - \zval_0^3)} \, .
\end{align}
And the second moment of $x$ is:
\begin{align}
    E[x^2] &= \frac{1}{V} \int_{0}^{2\pi} \int_{\zval_0}^{\zval_1} \int_{0}^{\baseradius} (rt\cos\theta)^2 \cdot r \zval^2 \, dr \, d\zval \, d\theta \\
    &= \frac{1}{V} \int_{\zval_0}^{\zval_1} \int_{0}^{\baseradius}  r^3 \zval^4 \int_{0}^{2\pi}  \cos^2\theta \, d\theta \, dr \, d\zval \\
    &= \frac{1}{V} \cdot \frac{\baseradius^4}{4} \cdot \frac{\zval_1^5 - \zval_0^5}{5} \cdot \pi \\
    &= \frac{\baseradius^2}{4} \cdot \frac{3(\zval_1^5 - \zval_0^5)}{5(\zval_1^3 - \zval_0^3)} \, .
\end{align}
The second moment of $y$ is the same by symmetry. All cross terms in the covariance are z, also by symmetry.

With these moments defined, we can construct the mean and covariance for a random point within our conical frustum. The mean along the ray direction $\mu_\zval$ is simply the first moment with respect to $\zval$:
\begin{equation}
\mu_\zval = \frac{3\left(\zval_1^4-\zval_0^4\right)}{4\left(\zval_1^3-\zval_0^3\right)}\, .
\end{equation}
The variance of the conical frustum with respect to $t$ follows from the definition of variance as $\operatorname{Var}(\zval)=\operatorname{E} \left[\zval^{2}\right]-\operatorname{E}[\zval]^{2}$:
\begin{equation}
    \sigma_\zval^2 = \frac{3 \left(\zval_1^5 - \zval_0^5\right)}{5 \left( \zval_1^3 - \zval_0^3\right)} - \mu_\zval^2\,.
\end{equation}
The variance of the conical frustum with respect to its radius $r$ is equal to the variance of the frustum with respect to $x$ or (by symmetry) $y$. Since the first moment with respect to $x$ is zero, the variance is equal to the second moment:
\begin{equation}
    \sigma_{r}^2 = \baseradius^2 \left(\frac{3 \left( \zval_1^5 - \zval_0^5 \right)}{20 \left(\zval_1^3 - \zval_0^3\right)}\right)\,.
\end{equation}
Computing all three of these quantities in their given form is numerically unstable --- the ratio of the differences between $\zval_1$ and $\zval_0$ raised to large powers is difficult to compute accurately when $\zval_0$ and $\zval_1$ are near each other, which occurs frequently during training. Using these quantities in practice often produces 0 or NaN instead of accurate values, which causes training to fail.
We therefore reparameterize these equations as a function of the center and spread of $\zval_0$ and $\zval_1$: $\zc = (\zval_0 + \zval_1) / 2$, $\zd = (\zval_1 - \zval_0) / 2$.
This allows us to rewrite each mean and variance as a first-order term that is then corrected by higher-order terms, which are scaled by $\zd$. This gives us stable and accurate values even when $\zd$ is small. Our reparameterized values are:
\begin{gather}
    \mu_\zval = \zc + \frac{2 \zc \zd^2}{3 \zc^2 + \zd^2}\,,  \quad\,\,
    \sigma_\zval^2 = \frac{\zd^2}{3} -\frac{4 \zd^4 (12 \zc^2 - \zd^2)}{15 (3 \zc^2 + \zd^2)^2}\,, \nonumber \\
    \sigma_{r}^2 = \baseradius^2 \left( \frac{\zc^2}{4} + \frac{5 \zd^2}{12} - \frac{4 \zd^4}{15 (3 \zc^2 + \zd^2)} \right) \,.
\end{gather}

Note that our multivariate Gaussian approximation of a conical frustum will be inaccurate if there is a significant difference between the base and top radii of the frustum, which will be true for frustums that are very near the camera's center of projection when the camera FOV is large. This is highly uncommon in most datasets, but may be an issue if one were to use \shortmodelnamelower in unusual circumstances, such as macro photography with a fisheye lens.

\section{The $L$ Hyperparameter in PE and IPE}
\label{sec:Lparam}

\begin{figure}[t]
    \centering
    \includegraphics[width=2.5in]{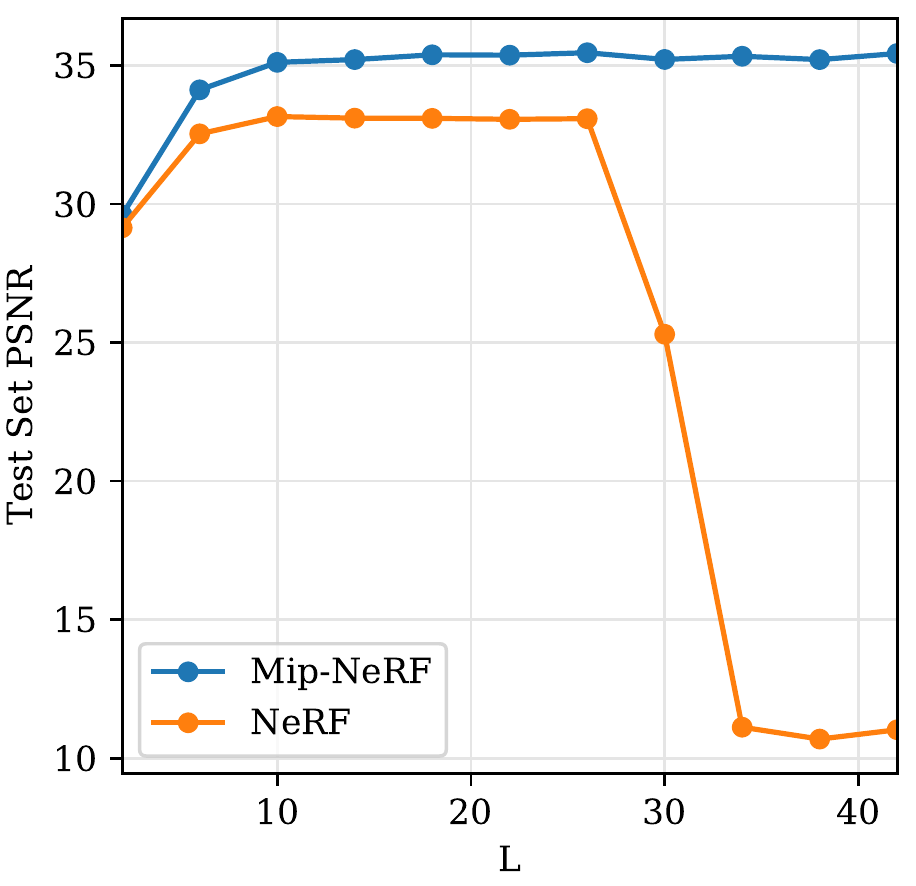}
    \caption{
    PSNRs for NeRF and \shortmodelnamelower on the test set of the \scenename{lego} scene, as we vary the positional encoding degree $L$. In NeRF, performance decreases due to overfitting for large values of $L$, but in \shortmodelnamelower this parameter is effectively removed from tuning --- it can just be set to a large value and forgotten, because IPE features ``tune'' their own frequencies automatically.
    }
    \label{fig:lsweep}
\end{figure}

IPE features can be viewed as a generalization of PE features: $\gamma(\position) = \gamma(\boldsymbol{\mu} = \position, \covmat = \mathbf{0})$. Or more rigorously, PE features can be thought of as ``hard'' IPE features in which all points are assumed to have identical isotropic covariance matrices whose variance has been heuristically determined by the $L$ hyperparameter: the value of $L$ determines the frequency at which PE features are truncated, just as the Gaussian function of variance in IPE serves as a ``soft'' truncation of IPE features. Because the ``soft'' maximum frequency of IPE features is determined entirely by the geometry and intrinsics of the camera, IPE features do not depend on the $L$ hyperparameter, and so using IPE features removes the need for tuning $L$.
This is because in PE the $L$ parameter determines where the high frequencies in the PE are truncated, but in IPE those high frequencies are naturally attenuated by the size of the multivariate Gaussian used as input to the encoding: the smaller the Gaussian, the more high frequencies will be retained. 
To demonstrate this, we performed as ``sweep'' of L in both \shortmodelnamelower and NeRF, and report the test-set PSNR for a single scene, which is visualized in Figure~\ref{fig:lsweep}. We see that in NeRF, there is a range of values for $L$ in which performance is maximized, but values that are too large or too small will hurt performance. But in \shortmodelnamelower, we see that $L$ can be set to an arbitrarily large value and performance is unaffected. In practice, in all \shortmodelnamelower experiments in the paper we set $L=16$, which is a value that results in the last dimension of all IPE features constructed during training to be less than numerical epsilon.

\section{Hyperparameters}

In all experiments in the paper we take care to use exactly the same set of hyperparameters that were used in Mildenhall~\etal~\cite{mildenhall2020}, so as to isolate the specific contributions of \shortmodelnamelower as they relate to cone-casting and IPE features. The three relevant hyperparameters that govern \shortmodelnamelower's behavior are: 1) the number of samples $N$ drawn at each of the two levels ($N=128$), 2) the histogram ``padding'' hyperparameter $\dirichlet$ on the coarse transmittance weights that are used to sample the fine $\zval$ values ($\dirichlet=0.01$), and 3) the multiplier $\lambda$ on the ``coarse'' component of the loss function ($\lambda=0.1$). And though \shortmodelnamelower adds these three hyperparameters, it also deprecates three NeRF hyperparameters that are no longer used: 1) The number of samples $N_c$ drawn for the ``coarse'' MLP ($N_c=64$), 2) The number of samples $N_f$ drawn for the ``fine'' MLP ($N_f=128$), and 3) The degree $L$ used for the spatial positional encoding ($L=10$). The $\dirichlet$ parameter used by \shortmodelnamelower serves a similar purpose as the balance between $N_c$ and $N_f$ did in NeRF --- a larger value of $\dirichlet$ biases the final samples used during rendering towards a uniform distribution, just as a larger value of $N_c$ biases the final samples (which are the sorted union of the uniform coarse samples and the biased fine samples) towards a uniform distribution. \shortmodelname's multiplier $\lambda$ has no analog in NeRF, as NeRF's usage of two distinct MLPs means that the ``coarse'' and ``fine'' losses in NeRF do not need to be balanced --- thankfully, though \shortmodelnamelower adds the need to tune this new hyperparameter $\lambda$, it simultaneously removes the need to tune the $L$ hyperparameter as discussed in Section~\ref{sec:Lparam}, so the total number of hyperparameters that require tuning remains constant across the two models.

Before running the experiments in the paper, we briefly tuned the $\dirichlet$ and $\lambda$ hyperparameters by hand on the validation set of the \scenename{lego} scene. $N$ was not tuned, and was just set to $128$ such that the total number of MLP evaluations used by \shortmodelnamelower matched the total number used by NeRF.

\section{Forward-Facing Scenes}

Note that this paper does not evaluate on the LLFF dataset~\cite{mildenhall2019}, which consists of scenes captured by a ``forward-facing'' handheld cellphone camera. For these scenes, NeRF trained and evaluated models in a ``normalized device coordinates'' (NDC) space. NDC coordinates work by nonlinearly warping a frustum-shaped space into a unit cube, which sidesteps some otherwise challenging design decisions (such as how an unbounded 3D space should be represented using positional encoding). NDC coordinates can only be used for these ``forward-facing'' scenes; in scenes where the camera rotates significantly (which is the case for the vast majority of 3D datasets) NeRF uses conventional 3D ``world coordinates''.
One interesting consequence of NDC space is that the 3D volume corresponding to a pixel is \emph{not} a frustum, but is instead a rectangle --- in NDC the spatial support of a pixel in the $xy$ plane \emph{does not} increase with the distance from the image plane, as it would in conventional projective geometry.

We briefly experimented with a variant of \shortmodelnamelower that works in NDC space by casting \emph{cylinders} instead of cones.
The average PSNR achieved by JaxNeRF on this task is $26.843$, and this cylinder-casting variant of \shortmodelnamelower achieves an average PSNR of $26.838$. Because this \shortmodelnamelower variant roughly matches the accuracy of NeRF, the only substantial benefit it appears to provide is removing the need to tune the $L$ parameter in positional encoding. This result provides some insight into why NeRF works so well on forward-facing scenes: in NDC space there is little difference between NeRF's ``incorrect'' aliased approach of casting rays and tuning the $L$ hyperparameter (which as discussed in Section~\ref{sec:Lparam}, is approximately equivalent to using IPE features with isotropic Gaussians) and the more ``correct'' anti-aliased approach of \shortmodelnamelower. In essence, NeRF is already able to get most of the benefit provided by cone-casting and IPE features in NDC space, because in NDC space NeRF's aliased model is already very similar to \shortmodelnamelower's approach. This interplay between scene parameterization and anti-aliasing suggests that a signal processing analysis of coordinate spaces in neural rendering problems may provide additional unexpected benefits or insights.

\section{Model Details}

The primary contributions of this paper are the use of cone tracing, integrated positional encoding features, and our use of a single unified multiscale model (as opposed to NeRF's separate per-scale models), which together allow \shortmodelnamelower to better handle multiscale data and reduce aliasing. Additionally, \shortmodelnamelower includes a small number of changes that do not meaningfully change \shortmodelnamelower's accuracy or speed, but slightly simplify our method and increase its robustness during optimization. These ``miscellaneous'' changes, as noted by the ``w/o Misc.'' ablation in the main paper, do not significantly affect \shortmodelnamelower's performance, but are described here in full for the sake of reproducibility with the hopes that future work will find them useful.

\subsection{Identity Concatenation}

In the original NeRF paper, the input to the MLP is not just the positional encoding of the position and view direction, but is instead the concatenation of the positional encoding with the position and view direction being encoded. We found this ``identity'' encoding to not contribute meaningfully to performance or speed, and its presence makes the formalization of our IPE features somewhat challenging, so this in \shortmodelnamelower this identity mapping is removed and the only input to the MLP is the integrated positional encoding itself.

\subsection{Activation Functions}

In the original NeRF paper, the activation functions used by the MLP to construct the predicted density $\density$ and color $\col$ are a ReLU and a sigmoid, respectively. Instead of a ReLU as the activation function to produce $\density$, we use a shifted softplus: $\log(1+\exp(x - 1))$. We found that using a softplus yielded a smoother optimization problem that is less prone to catastrophic failure modes in which the MLP emits negative values everywhere (in which case all gradients from $\density$ are zero and optimization will fail). The shift by $-1$ within the softplus is equivalent to initializing the biases that produce $\density$ in \shortmodelnamelower to $-1$, and this causes initial $\density$ values to be small. Initializing the density of the NeRF to small values results in slightly faster optimization at the beginning of training, as dense scene content causes gradients from scene content ``behind'' that dense content to be suppressed.
Instead of a sigmoid to produce color $\col$, we use a ``widened'' sigmoid that saturates slightly outside of $[0, 1]$ (the range of input RGB intensities): $(1 + 2\epsilon) / (1+\exp(-x)) - \epsilon$, with $\epsilon=0.001$. This avoids an uncommon failure mode in which training tries to explain away a black or white pixel by saturating network activations into the tails of the sigmoid where the gradient is zero, which may cause optimization to fail. By having the network saturate at values slightly outside of the range of input values, activations are never encouraged to saturate.
These changes to activation functions have little effect on performance, but we found that they improved training stability when using large learning rates (though all results in this paper use the same lower learning rates used by Mildenhall~\etal~\cite{mildenhall2020} for fair comparison).

\begin{table*}[]
    \centering
    \resizebox{\linewidth}{!}{
    \huge
    \begin{tabular}{@{}l@{\,\,}|c@{\,\,\,}c@{\,\,\,}c@{\,\,\,}c|c@{\,\,\,}c@{\,\,\,}c@{\,\,\,}c|c@{\,\,\,}c@{\,\,\,}c@{\,\,\,}c|c|c@{\,\,\,}c@{\,\,\,}c@{}}
    & \multicolumn{4}{c|}{PSNR $\uparrow$} & \multicolumn{4}{c|}{SSIM $\uparrow$} & \multicolumn{4}{c|}{LPIPS $\downarrow$} & & Train Time & Test Time & \\
    & Full Res. & $\nicefrac{1}{2}$ Res. & $\nicefrac{1}{4}$ Res. & $\nicefrac{1}{8}$ Res. & Full Res. & $\nicefrac{1}{2}$ Res. & $\nicefrac{1}{4}$ Res. & $\nicefrac{1}{8}$ Res. & Full Res. & $\nicefrac{1}{2}$ Res. & $\nicefrac{1}{4}$ Res. & $\nicefrac{1}{8}$ Res & Avg. $\downarrow$ & (hours) & (sec/MP) & \# Params \\ \hline
NeRF + Area, Center, 1$\times$ SS  &                    27.471 &                    28.016 &                    27.816 &                    26.657 &                    0.9187 &                    0.9301 &                    0.9365 &                    0.9304 &                    0.1064 &                    0.0924 &                    0.0934 &                    0.1064 &                    0.0362 & 2.85 & 2.61 & 1,191\thousand \\
NeRF + Area, Center, 4$\times$ SS  & \cellcolor{yellow} 28.424 & \cellcolor{yellow} 29.420 & \cellcolor{yellow} 29.863 & \cellcolor{yellow} 29.233 & \cellcolor{yellow} 0.9297 & \cellcolor{yellow} 0.9426 & \cellcolor{yellow} 0.9526 & \cellcolor{yellow} 0.9547 & \cellcolor{yellow} 0.0807 & \cellcolor{yellow} 0.0598 & \cellcolor{yellow} 0.0530 & \cellcolor{yellow} 0.0536 & \cellcolor{yellow} 0.0259 & 17.69 & 10.44 & 1,191\thousand \\
NeRF + Area, Center, 16$\times$ SS & \cellcolor{orange} 31.566 & \cellcolor{orange} 33.116 & \cellcolor{orange} 33.982 & \cellcolor{orange} 32.933 & \cellcolor{orange} 0.9524 & \cellcolor{orange} 0.9660 & \cellcolor{orange} 0.9753 & \cellcolor{orange} 0.9768 & \cellcolor{orange} 0.0537 & \cellcolor{orange} 0.0316 & \cellcolor{orange} 0.0227 & \cellcolor{orange} 0.0216 & \cellcolor{orange} 0.0144 & 37.18 & 41.76 & 1,191\thousand \\
\hline
\shortmodelname                    & \cellcolor{red}    32.629 & \cellcolor{red}    34.336 & \cellcolor{red}    35.471 & \cellcolor{red}    35.602 & \cellcolor{red}    0.9579 & \cellcolor{red}    0.9703 & \cellcolor{red}    0.9786 & \cellcolor{red}    0.9833 & \cellcolor{red}    0.0469 & \cellcolor{red}    0.0260 & \cellcolor{red}    0.0168 & \cellcolor{red}    0.0120 & \cellcolor{red}    0.0114 & 2.79 & 2.48 & 612\thousand \\
    \end{tabular}
    }
    \caption{
    Here we evaluate \shortmodelnamelower against an extension of NeRF in which brute-force supersampling with jittered rays is used during training and evaluation, on our multiscale Blender dataset (``16$\times$ SS'' indicates that 16 rays are cast per pixel, etc). \shortmodelname is able to outperform this baseline by a significant margin in terms of quality, while also being $13 \times$ faster to train and $16 \times$ faster to evaluate.}
    \label{tab:supertrain}
\end{table*}

\subsection{Optimization}

\begin{figure}[t]
    \centering
    \includegraphics[width=2.5in]{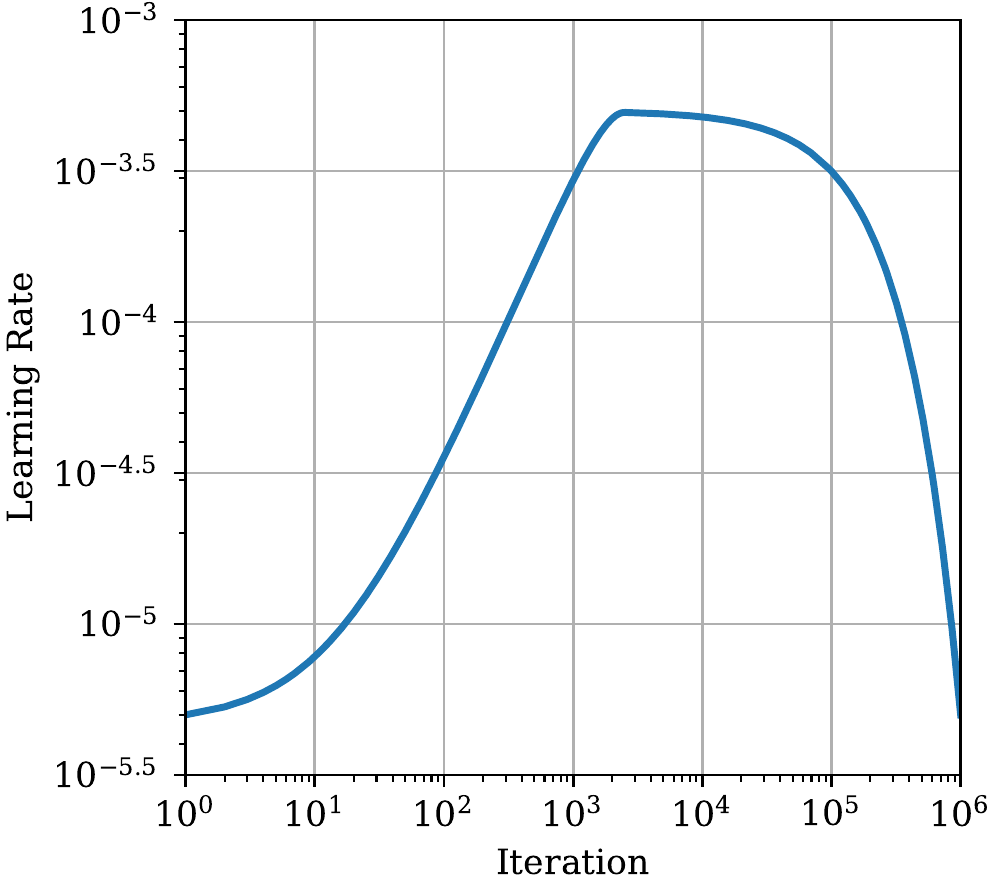}
    \caption{
    The learning rate used in all JaxNeRF and \shortmodelnamelower experiments.
    }
    \label{fig:lr}
\end{figure}

In all experiments we train \shortmodelnamelower and JaxNeRF using the default training procedure specified in the JaxNeRF codebase: 1 million iterations of Adam~\cite{adam} with a batch size of $4096$ and a learning rate that is annealed logarithmically from $\learnrate_{0} = 5 \cdot 10^{-4}$ to $\learnrate_{n} =5 \cdot 10^{-6}$. We additionally ``warm up'' the learning rate using the functionality provided by JaxNeRF, which does not improve the performance of \shortmodelnamelower itself, but which we found to improve the stability of some of the \shortmodelnamelower ablations. To allow our ablations to be competitive, and to enable a fair comparison across all models, we therefore use this warm up strategy in all \shortmodelnamelower and JaxNeRF experiments. Because the warm up procedure in JaxNeRF is not described in its documentation~\cite{jaxnerf2020github}, for the sake of reproducibility we will describe it here. For the first $\numsteps_{w} = 2500$ iterations of optimization, we scale the basic learning rate by an additional scale factor that is smoothly annealed between $\lambda_{w} = 0.01$ and $1$ during this warm up period.
The learning rate at iteration $i$ during training is:
\begin{align}
\learnrate_i = & \left( \lambda_{w} + (1 - \lambda_{w}) \sin(
        (\nicefrac{\pi}{2}) \operatorname{clip}(\nicefrac{i}{\numsteps_{w}}, 0, 1)) \right) \nonumber \\
        & \times \left( \exp((1 - \nicefrac{i}{\numsteps})\log(\learnrate_{0}) + (\nicefrac{i}{\numsteps}) \log(\learnrate_{n})) \right)
\end{align}
See Figure~\ref{fig:lr} for a visualization.

\subsection{View Dependent Effects}

We handle viewing directions exactly as was done in NeRF: the ray direction $\raydir$ is normalized, positionally encoded ($L=4$), and injected into the last layer of the MLP after $\density$ is predicted but before $\col$ is predicted. This is omitted from our notation in the main paper for simplicity's sake. see Mildenhall~\etal for details~\cite{mildenhall2020}.

\section{Supersampling Baseline}

In the main paper we presented a generous baseline approach in which NeRF is trained on only full-resolution images (thereby sidestepping its poor performance when trained on multi-resolution data) and then evaluated on our multiscale Blender dataset by brute-force supersampling: rendering a full-resolution image that is then downsampled to match the resolution of the ground truth. This roughly matches the performance of \shortmodelnamelower, but is $22 \times$ slower and relies on ``oracle'' scale information that does not exist for most datasets. Here we explore an alternative supersampling baseline, in which we train an extension of NeRF on the multiscale dataset while supersampling during \emph{both} training and evaluation: for every pixel we cast multiple jittered rays (sampled uniformly at random) through the spatial footprint of each pixel, render each ray with the NeRF, and then use the mean of those rendered values as the predicted color of that pixel in the loss function. As shown by the results of this experiment (Table~\ref{tab:supertrain}) this brute-force supersampling model not only performs worse than \shortmodelnamelower even when casting as many as $16$ rays per pixel, but is also significantly more expensive during both training and evaluation.

\section{Alternative Gaussian Positional Encoding}

During experimentation we explored alternative approaches for featurizing the mean and covariance matrix of the multivariate Gaussians used by \shortmodelnamelower. One such alternative strategy is to simply apply positional encoding to the mean and to the (signed) square root of the elements of the covariance matrix, and use the concatenation of the two as input. Specifically, we compute the positional encoding of $\boldsymbol{\mu}$ with $L=12$, and compute the positional encoding of $\operatorname{vec}(\operatorname{triu}(\operatorname{sign}(\covmat) \circ \sqrt{|\covmat|}))$ with $L=2$. We found that this approach performs comparably to the IPE features presented in the main paper, as shown in Table~\ref{tab:alt_encoding}. We chose to advocate for IPE features in the main paper instead of this concatenation alternative because 1)  IPE features are more compact (thereby reducing model size and evaluation time), 2) IPE features are easy to justify and reason about (as they approximate an expectation of positional encoding features with respect to a conical frustum), and 3) IPE features have no hyperparameters (while this concatenation alternative is sensitive to its two $L$ hyperparameters and the design decisions used when parameterizing $\covmat$).

This experiment with using this alternative to IPE also provides some insight into the inner workings of \shortmodelnamelower. While IPE features are insensitive to the off-diagonal elements of $\covmat$, this concatenation alternative should endow the MLP with the ability to reason about the correlation of dimensions of the multivariate Gaussian. The fact that this ability does not improve accuracy may suggest that correlation is not a helpful cue, which contradicted the intuition of the authors. 
Additionally, this experiment reinforces the assertions made in the paper that the reason for \shortmodelnamelower's improved performance is its explicit modeling of conical frustums, as opposed to NeRF's usage of point samples along a ray. Though it is critical that the geometry of image formation be modeled accurately, there are likely many effective ways to featurize that geometry.

\begin{table}[]
    \centering
    \resizebox{\linewidth}{!}{
    \begin{tabular}{l|ccc|c}
    Multiscale Blender & PSNR $\uparrow$ & SSIM $\uparrow$ & LPIPS $\downarrow$ & Avg. $\downarrow$ \\ \hline
    Integrated PE        & \bf 34.51 &  \bf 0.973 & \bf 0.025 & \bf 0.0113 \\
    Concatenated PE & 34.40 &  \bf 0.973 & \bf 0.025 & 0.0114 \\
    \multicolumn{5}{c}{} \\
    Blender & PSNR $\uparrow$ & SSIM $\uparrow$ & LPIPS $\downarrow$ & Avg. $\downarrow$ \\ \hline
    Integrated PE        & \bf 33.09 & \bf 0.961 & 0.043 & 0.0161  \\
    Concatenated PE & \bf 33.09 & \bf 0.961 & \bf 0.042 & \bf 0.0160 \\
    \end{tabular}
    }
    \caption{
    An evaluation of the IPE features against an alternative approach in which the mean and covariance of the multivariate Gaussian corresponding to a conical frustum are positionally encoded and concatenated. Both approaches perform comparably on the multiscale and single-scale Blender datasets.
    }
    \label{tab:alt_encoding}
\end{table}

\section{Additional Results}

\begin{figure*}[t]
    \centering
    \begin{tabular}{@{}c@{\,\,}c|c|c|c@{}}
    \flatim{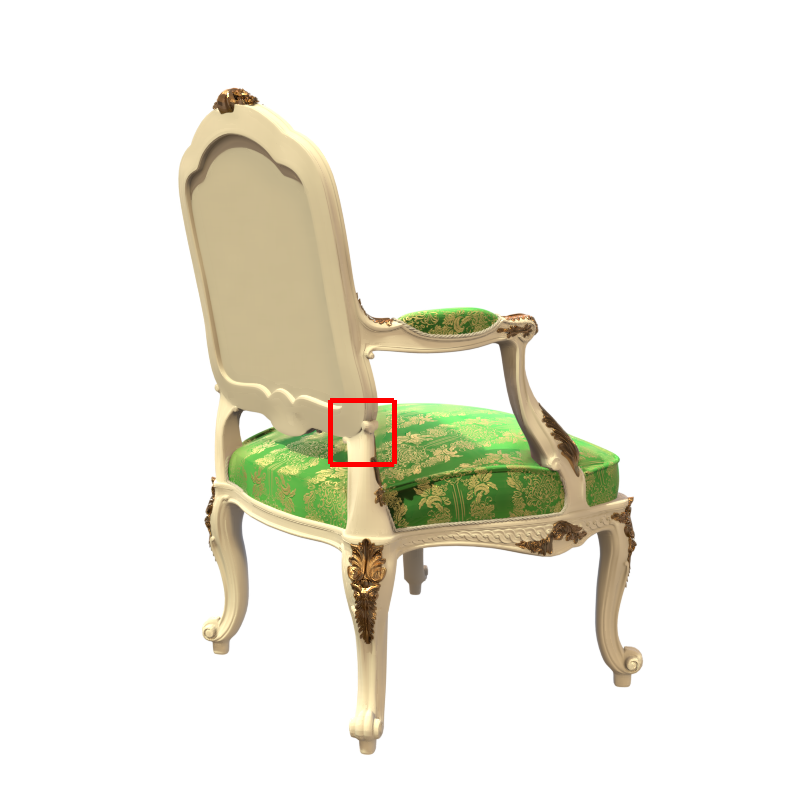} & 
    \textpyr{gt_chair_115_400_330_64.png}{}{}{}{} & 
    \textpyr{jaxnerf_multiblender_chair_115_400_330_64.png}{0.758}{0.857}{0.904}{0.850} & 
    \textpyr{jaxnerf_multiblender_extras_chair_115_400_330_64.png}{0.844}{0.918}{0.964}{0.941} & 
    \textpyr{prenerf_multiblender_chair_115_400_330_64.png}{ \color{wincolor} 0.863}{ \color{wincolor} 0.923}{ \color{wincolor} 0.970}{ \color{wincolor} 0.967} \\ 
    \flatim{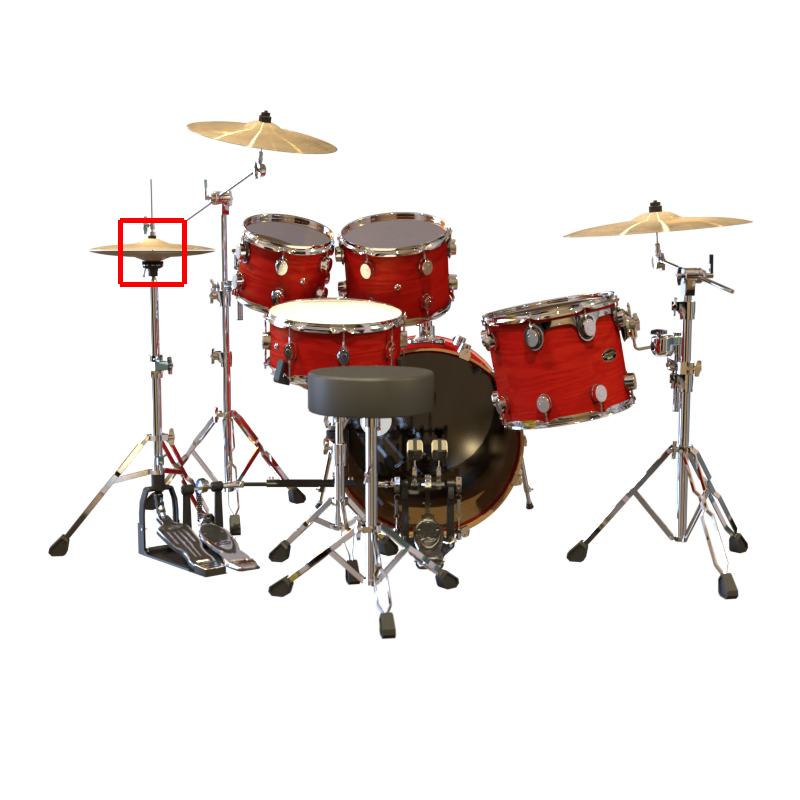} & 
    \textpyr{gt_drums_100_220_120_64.png}{}{}{}{} & 
    \textpyr{jaxnerf_multiblender_drums_100_220_120_64.png}{0.665}{0.804}{ \color{wincolor} 0.918}{0.814} & 
    \textpyr{jaxnerf_multiblender_extras_drums_100_220_120_64.png}{0.721}{0.825}{0.913}{0.896} & 
    \textpyr{prenerf_multiblender_drums_100_220_120_64.png}{ \color{wincolor} 0.771}{ \color{wincolor} 0.846}{0.915}{ \color{wincolor} 0.910} \\ 
    \flatim{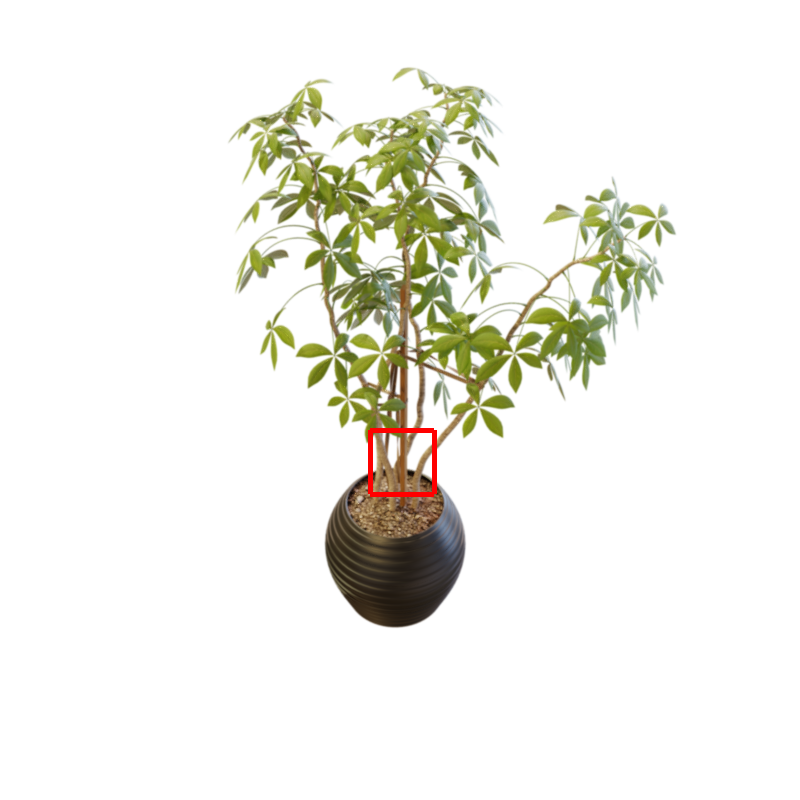} & 
    \textpyr{gt_ficus_30_430_370_64.png}{}{}{}{} & 
    \textpyr{jaxnerf_multiblender_ficus_30_430_370_64.png}{0.685}{0.832}{0.874}{0.779} & 
    \textpyr{jaxnerf_multiblender_extras_ficus_30_430_370_64.png}{0.868}{0.949}{0.971}{0.929} & 
    \textpyr{prenerf_multiblender_ficus_30_430_370_64.png}{ \color{wincolor} 0.922}{ \color{wincolor} 0.960}{ \color{wincolor} 0.973}{ \color{wincolor} 0.971} \\ 
    \flatim{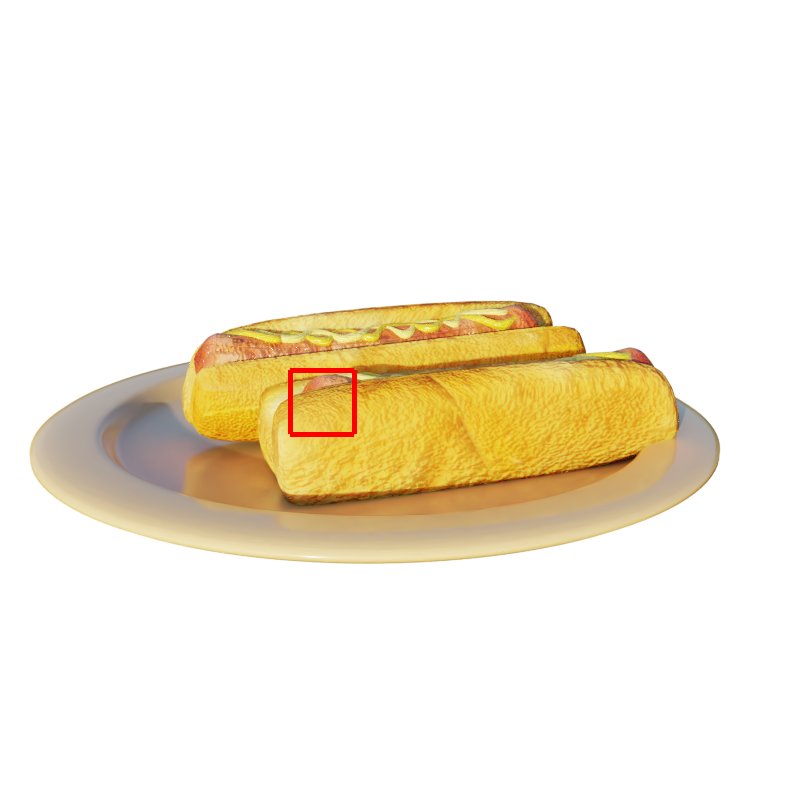} & 
    \textpyr{gt_hotdog_70_370_290_64.png}{}{}{}{} & 
    \textpyr{jaxnerf_multiblender_hotdog_70_370_290_64.png}{0.659}{0.794}{0.890}{0.763} & 
    \textpyr{jaxnerf_multiblender_extras_hotdog_70_370_290_64.png}{0.743}{0.863}{0.942}{0.906} & 
    \textpyr{prenerf_multiblender_hotdog_70_370_290_64.png}{ \color{wincolor} 0.864}{ \color{wincolor} 0.935}{ \color{wincolor} 0.965}{ \color{wincolor} 0.966} \\ 
    \flatim{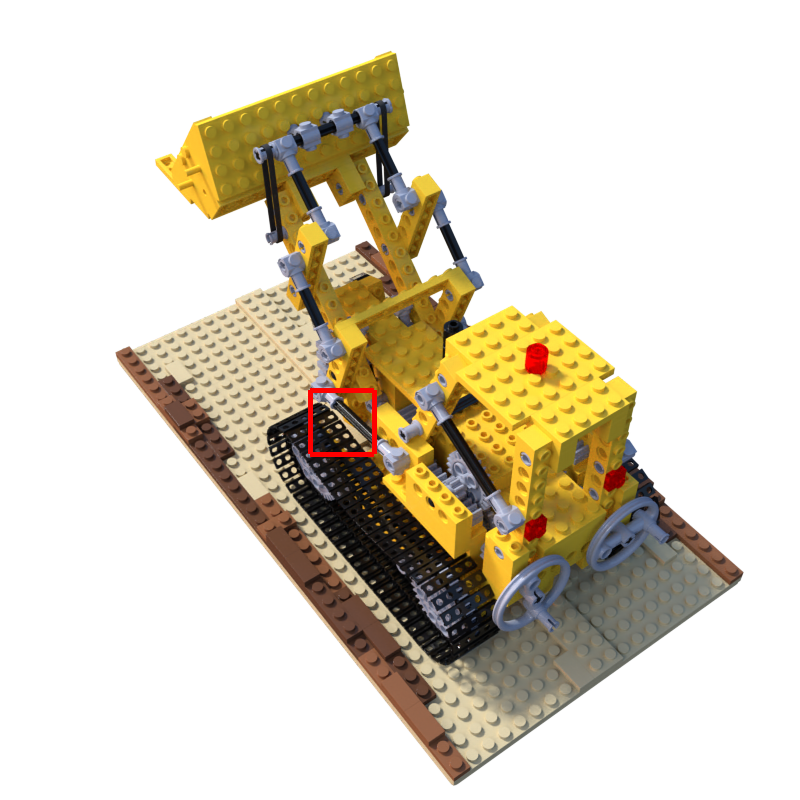} & 
    \textpyr{gt_lego_190_390_310_64.png}{}{}{}{} & 
    \textpyr{jaxnerf_multiblender_lego_190_390_310_64.png}{0.542}{0.668}{0.701}{0.734} & 
    \textpyr{jaxnerf_multiblender_extras_lego_190_390_310_64.png}{0.720}{0.843}{0.910}{0.747} & 
    \textpyr{prenerf_multiblender_lego_190_390_310_64.png}{ \color{wincolor} 0.824}{ \color{wincolor} 0.901}{ \color{wincolor} 0.940}{ \color{wincolor} 0.923} \\ 
    \flatim{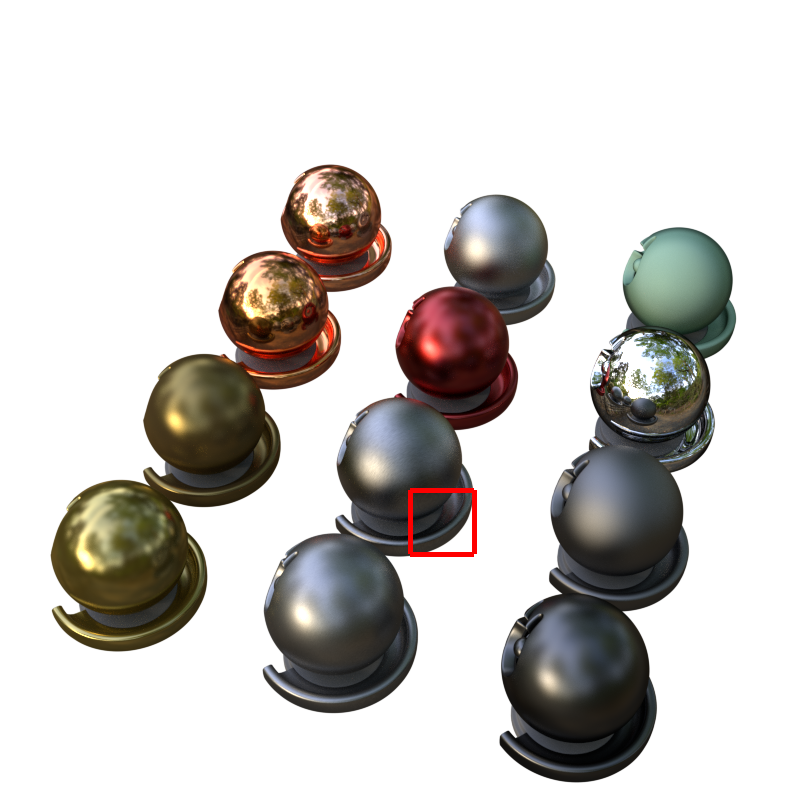} & 
    \textpyr{gt_materials_180_490_410_64.png}{}{}{}{} & 
    \textpyr{jaxnerf_multiblender_materials_180_490_410_64.png}{0.733}{0.832}{0.838}{0.787} & 
    \textpyr{jaxnerf_multiblender_extras_materials_180_490_410_64.png}{0.879}{0.926}{0.946}{0.971} & 
    \textpyr{prenerf_multiblender_materials_180_490_410_64.png}{ \color{wincolor} 0.935}{ \color{wincolor} 0.964}{ \color{wincolor} 0.975}{ \color{wincolor} 0.994} \\ 
    \flatim{gt_mic_120_entire.png} & 
    \textpyr{gt_mic_120_80_560_64.png}{}{}{}{} & 
    \textpyr{jaxnerf_multiblender_mic_120_80_560_64.png}{0.448}{0.562}{0.696}{0.906} & 
    \textpyr{jaxnerf_multiblender_extras_mic_120_80_560_64.png}{0.525}{0.633}{0.794}{0.918} & 
    \textpyr{prenerf_multiblender_mic_120_80_560_64.png}{ \color{wincolor} 0.785}{ \color{wincolor} 0.837}{ \color{wincolor} 0.861}{ \color{wincolor} 0.975} \\ 
    \flatim{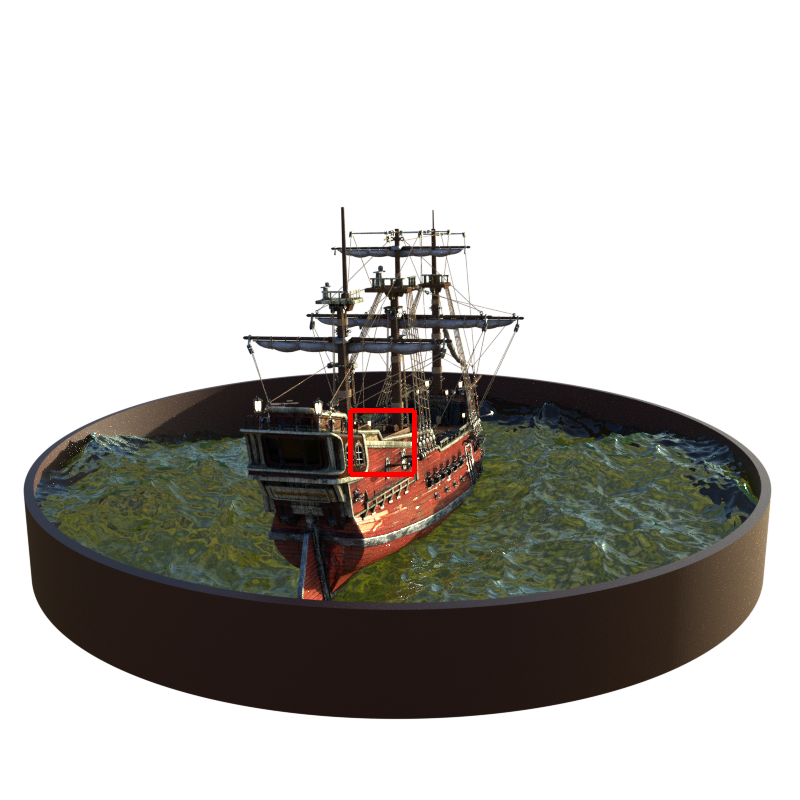} & 
    \textpyr{gt_ship_130_410_350_64.png}{}{}{}{} & 
    \textpyr{jaxnerf_multiblender_ship_130_410_350_64.png}{0.494}{0.742}{0.839}{0.810} & 
    \textpyr{jaxnerf_multiblender_extras_ship_130_410_350_64.png}{0.655}{0.859}{0.944}{0.833} & 
    \textpyr{prenerf_multiblender_ship_130_410_350_64.png}{ \color{wincolor} 0.790}{ \color{wincolor} 0.915}{ \color{wincolor} 0.967}{ \color{wincolor} 0.972} \\ 
    \multicolumn{2}{c|}{\footnotesize{Ground-Truth}} &
    \footnotesize{NeRF} & 
    \footnotesize{NeRF + Area, Center, Misc} & 
    \footnotesize{\shortmodelname}
    \end{tabular}
    \caption{
    Additional visualizations of the output renderings from \shortmodelnamelower compared to the ground truth, NeRF, and an improved version of NeRF presented in this work, on test set images from the 8 scenes in our multiscale Blender dataset, in the same format as Figure~\ref{fig:multi_vis}.
    }
    \label{fig:multi_vis2}
\end{figure*}

\noindent {\bf Multiscale Blender Dataset.} To demonstrate the relative accuracy of \shortmodelnamelower compared to NeRF on each individual scene in the multiscale Blender dataset, the error metrics for each individual scene are provided in Table~\ref{tab:avg_multiblender_perscene}. \shortmodelname yields a significant reduction in error compared to NeRF across all scenes. Renderings produced by \shortmodelnamelower and baseline algorithms compared to the ground truth can be visually inspected in Figures~\ref{fig:multi_vis} and \ref{fig:multi_vis2}.

\noindent {\bf Blender Dataset.} Test-set error metrics for each individual scene in the (single scale) Blender dataset of Mildenhall \etal~\cite{mildenhall2020} can be seen in Table~\ref{tab:blender_results}. \shortmodelname yields lower error rates than NeRF on all scenes and all metrics.

\begin{table*}[]
    \centering
    \small
    \begin{tabular}{l|cccccccc}
     & \multicolumn{8}{c}{Average PSNR} \\
 & \scenename{chair}  & \scenename{drums}  & \scenename{ficus}  & \scenename{hotdog}  & \scenename{lego}  & \scenename{materials}  & \scenename{mic}  & \scenename{ship} \\ \hline 
NeRF (Jax Implementation)~\cite{jaxnerf2020github,mildenhall2020} &                    29.923  &                    23.273  &                    27.153  &                    32.001  &                    27.748  &                    26.295  &                    28.401  &                    26.462  \\
NeRF + Area Loss                                                  &                    30.277  &                    24.032  &                    27.149  &                    32.025  &                    27.602  &                    26.533  &                    28.120  &                    26.834  \\
NeRF + Area, Centered Pixels                                      &                    33.460  &                    25.802  &                    30.400  &                    35.672  &                    31.606  &                    30.155  &                    32.633  &                    30.019  \\
NeRF + Area, Center, Misc.                                        &                    33.394  &                    25.874  &                    30.369  &                    35.641  &                    31.646  &                    30.184  &                    32.601  &                    30.092  \\
\hline
\shortmodelname                                                   & \cellcolor{yellow} 37.141  & \cellcolor{red}    27.021  & \cellcolor{red}    33.188  & \cellcolor{yellow} 39.313  & \cellcolor{orange} 35.736  & \cellcolor{yellow} 32.558  & \cellcolor{red}    38.036  & \cellcolor{red}    33.083  \\
\shortmodelname w/o Misc.                                         & \cellcolor{orange} 37.275  & \cellcolor{orange} 26.979  & \cellcolor{orange} 33.160  & \cellcolor{orange} 39.357  & \cellcolor{red}    35.749  & \cellcolor{orange} 32.563  & \cellcolor{yellow} 37.997  & \cellcolor{orange} 33.078  \\
\shortmodelname w/o Single MLP                                    & \cellcolor{red}    37.310  & \cellcolor{yellow} 26.922  & \cellcolor{yellow} 33.045  & \cellcolor{red}    39.378  & \cellcolor{yellow} 35.605  & \cellcolor{red}    32.635  & \cellcolor{orange} 38.016  & \cellcolor{yellow} 33.011  \\
\shortmodelname w/o Area Loss                                     &                    35.188  &                    26.063  &                    32.542  &                    37.165  &                    34.319  &                    31.004  &                    35.922  &                    31.636  \\
\shortmodelname w/o \shortfeaturename                             &                    33.559  &                    25.864  &                    30.499  &                    35.793  &                    31.728  &                    30.272  &                    32.736  &                    30.276  \\
\multicolumn{9}{c}{} \\
 & \multicolumn{8}{c}{Average SSIM} \\
 & \scenename{chair}  & \scenename{drums}  & \scenename{ficus}  & \scenename{hotdog}  & \scenename{lego}  & \scenename{materials}  & \scenename{mic}  & \scenename{ship} \\ \hline 
NeRF (Jax Implementation)~\cite{jaxnerf2020github,mildenhall2020} &                    0.9436  &                    0.8908  &                    0.9423  &                    0.9586  &                    0.9256  &                    0.9335  &                    0.9580  &                    0.8607  \\
NeRF + Area Loss                                                  &                    0.9488  &                    0.9028  &                    0.9429  &                    0.9622  &                    0.9274  &                    0.9372  &                    0.9592  &                    0.8610  \\
NeRF + Area, Centered Pixels                                      &                    0.9710  &                    0.9310  &                    0.9705  &                    0.9794  &                    0.9643  &                    0.9670  &                    0.9800  &                    0.8994  \\
NeRF + Area, Center, Misc.                                        &                    0.9707  &                    0.9318  &                    0.9705  &                    0.9793  &                    0.9646  &                    0.9671  &                    0.9799  &                    0.9004  \\
\hline
\shortmodelname                                                   & \cellcolor{orange} 0.9875  & \cellcolor{red}    0.9450  & \cellcolor{red}    0.9836  & \cellcolor{red}    0.9880  & \cellcolor{red}    0.9843  & \cellcolor{red}    0.9767  & \cellcolor{red}    0.9928  & \cellcolor{orange} 0.9221  \\
\shortmodelname w/o Misc.                                         & \cellcolor{red}    0.9877  & \cellcolor{orange} 0.9448  & \cellcolor{orange} 0.9835  & \cellcolor{red}    0.9880  & \cellcolor{orange} 0.9842  & \cellcolor{red}    0.9767  & \cellcolor{orange} 0.9927  & \cellcolor{red}    0.9227  \\
\shortmodelname w/o Single MLP                                    & \cellcolor{orange} 0.9875  & \cellcolor{yellow} 0.9432  & \cellcolor{yellow} 0.9829  & \cellcolor{orange} 0.9876  & \cellcolor{yellow} 0.9836  & \cellcolor{orange} 0.9763  & \cellcolor{yellow} 0.9922  & \cellcolor{yellow} 0.9211  \\
\shortmodelname w/o Area Loss                                     & \cellcolor{yellow} 0.9817  &                    0.9371  &                    0.9823  & \cellcolor{yellow} 0.9849  &                    0.9792  & \cellcolor{yellow} 0.9731  &                    0.9911  &                    0.9175  \\
\shortmodelname w/o \shortfeaturename                             &                    0.9714  &                    0.9322  &                    0.9713  &                    0.9796  &                    0.9658  &                    0.9678  &                    0.9804  &                    0.9039  \\
\multicolumn{9}{c}{} \\
 & \multicolumn{8}{c}{Average LPIPS} \\
 & \scenename{chair}  & \scenename{drums}  & \scenename{ficus}  & \scenename{hotdog}  & \scenename{lego}  & \scenename{materials}  & \scenename{mic}  & \scenename{ship} \\ \hline 
NeRF (Jax Implementation)~\cite{jaxnerf2020github,mildenhall2020} &                    0.0347  &                    0.0689  &                    0.0324  &                    0.0279  &                    0.0410  &                    0.0452  &                    0.0307  &                    0.0948  \\
NeRF + Area Loss                                                  &                    0.0414  &                    0.0762  &                    0.0438  &                    0.0365  &                    0.0568  &                    0.0499  &                    0.0444  &                    0.1139  \\
NeRF + Area, Centered Pixels                                      &                    0.0281  &                    0.0593  &                    0.0264  &                    0.0240  &                    0.0348  &                    0.0330  &                    0.0249  &                    0.0865  \\
NeRF + Area, Center, Misc.                                        &                    0.0283  &                    0.0586  &                    0.0264  &                    0.0241  &                    0.0346  &                    0.0330  &                    0.0249  &                    0.0850  \\
\hline
\shortmodelname                                                   & \cellcolor{red}    0.0111  & \cellcolor{orange} 0.0439  & \cellcolor{red}    0.0135  & \cellcolor{red}    0.0121  & \cellcolor{red}    0.0127  & \cellcolor{red}    0.0186  & \cellcolor{red}    0.0065  & \cellcolor{orange} 0.0624  \\
\shortmodelname w/o Misc.                                         & \cellcolor{red}    0.0111  & \cellcolor{red}    0.0436  & \cellcolor{orange} 0.0136  & \cellcolor{yellow} 0.0123  & \cellcolor{red}    0.0127  & \cellcolor{red}    0.0186  & \cellcolor{orange} 0.0066  & \cellcolor{red}    0.0620  \\
\shortmodelname w/o Single MLP                                    & \cellcolor{orange} 0.0113  & \cellcolor{yellow} 0.0443  & \cellcolor{yellow} 0.0142  & \cellcolor{orange} 0.0122  & \cellcolor{orange} 0.0132  & \cellcolor{orange} 0.0187  & \cellcolor{yellow} 0.0068  & \cellcolor{yellow} 0.0628  \\
\shortmodelname w/o Area Loss                                     & \cellcolor{yellow} 0.0171  &                    0.0503  &                    0.0146  &                    0.0151  & \cellcolor{yellow} 0.0163  & \cellcolor{yellow} 0.0259  &                    0.0095  &                    0.0665  \\
\shortmodelname w/o \shortfeaturename                             &                    0.0276  &                    0.0578  &                    0.0259  &                    0.0240  &                    0.0340  &                    0.0320  &                    0.0231  &                    0.0829  \\
\multicolumn{9}{c}{} \\

    \end{tabular}
    \caption{Per-scene results on the test set images of the multiscale Blender dataset presented in this work.
    We report the arithmetic mean of each metric averaged over the four scales used in the dataset.
    }
    \label{tab:avg_multiblender_perscene}
\end{table*}

\begin{figure*}[t]
    \centering
    \begin{tabular}{@{}c@{\,\,}c|c|c|c@{}}
    \flatim{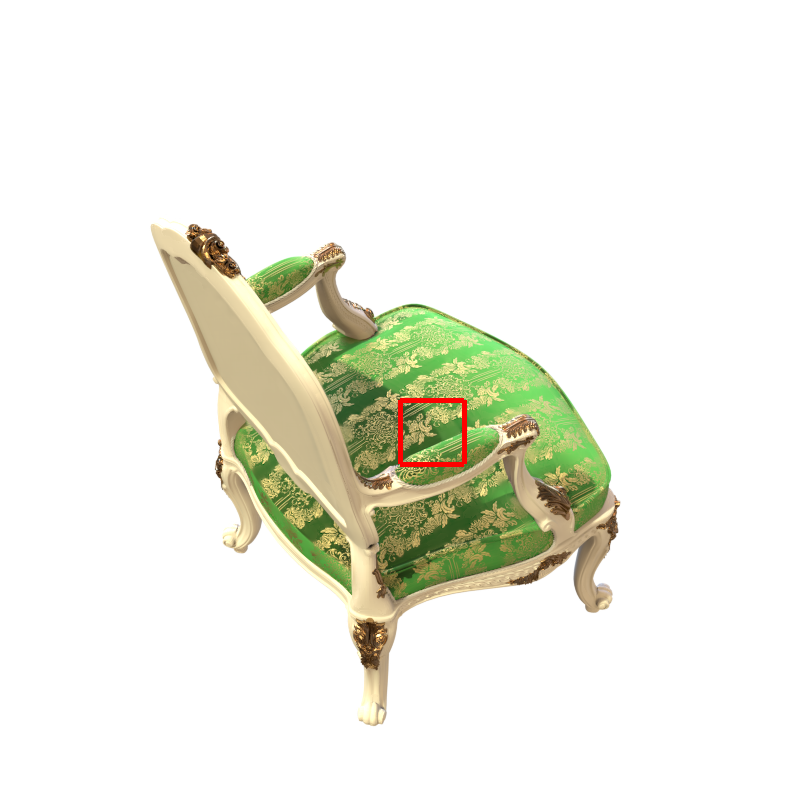} & 
    \textpyr{gt_chair_15_400_400_64.png}{}{}{}{} & 
    \textpyr{jaxnerf_multiblender_chair_15_400_400_64.png}{0.531}{0.798}{0.851}{0.690} & 
    \textpyr{jaxnerf_multiblender_extras_chair_15_400_400_64.png}{0.709}{0.925}{0.949}{0.921} & 
    \textpyr{prenerf_multiblender_chair_15_400_400_64.png}{ \color{wincolor} 0.907}{ \color{wincolor} 0.959}{ \color{wincolor} 0.973}{ \color{wincolor} 0.978} \\ 
    \flatim{gt_drums_0_entire.png} & 
    \textpyr{gt_drums_0_440_150_64.png}{}{}{}{} & 
    \textpyr{jaxnerf_multiblender_drums_0_440_150_64.png}{0.709}{0.910}{0.931}{0.663} & 
    \textpyr{jaxnerf_multiblender_extras_drums_0_440_150_64.png}{0.863}{0.959}{0.971}{0.881} & 
    \textpyr{prenerf_multiblender_drums_0_440_150_64.png}{ \color{wincolor} 0.940}{ \color{wincolor} 0.979}{ \color{wincolor} 0.989}{ \color{wincolor} 0.978} \\ 
    \flatim{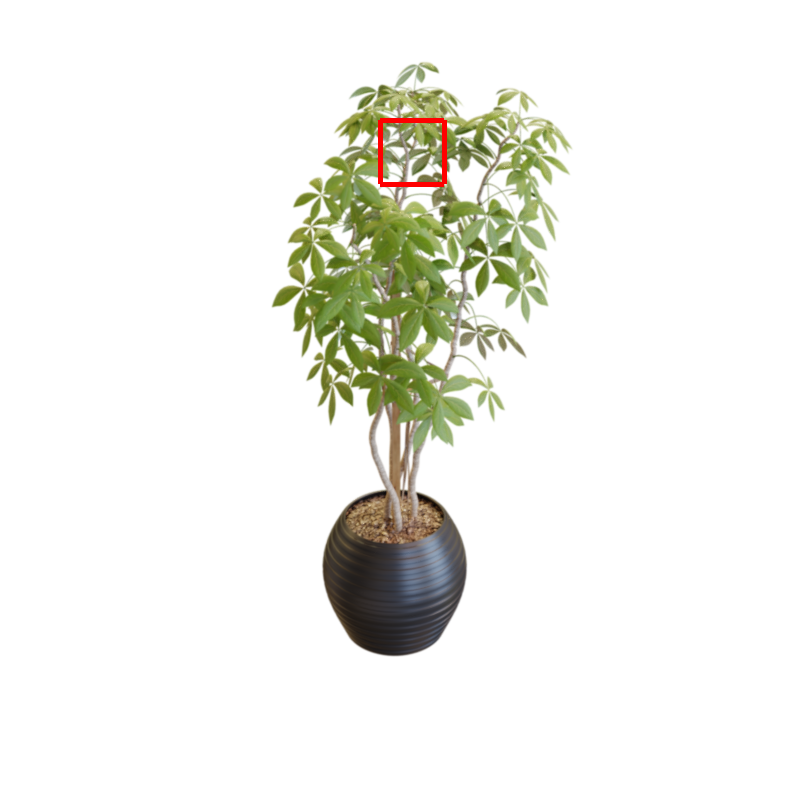} & 
    \textpyr{gt_ficus_150_120_380_64.png}{}{}{}{} & 
    \textpyr{jaxnerf_multiblender_ficus_150_120_380_64.png}{0.565}{0.717}{0.661}{0.258} & 
    \textpyr{jaxnerf_multiblender_extras_ficus_150_120_380_64.png}{0.710}{0.850}{0.916}{0.665} & 
    \textpyr{prenerf_multiblender_ficus_150_120_380_64.png}{ \color{wincolor} 0.850}{ \color{wincolor} 0.928}{ \color{wincolor} 0.944}{ \color{wincolor} 0.878} \\ 
    \flatim{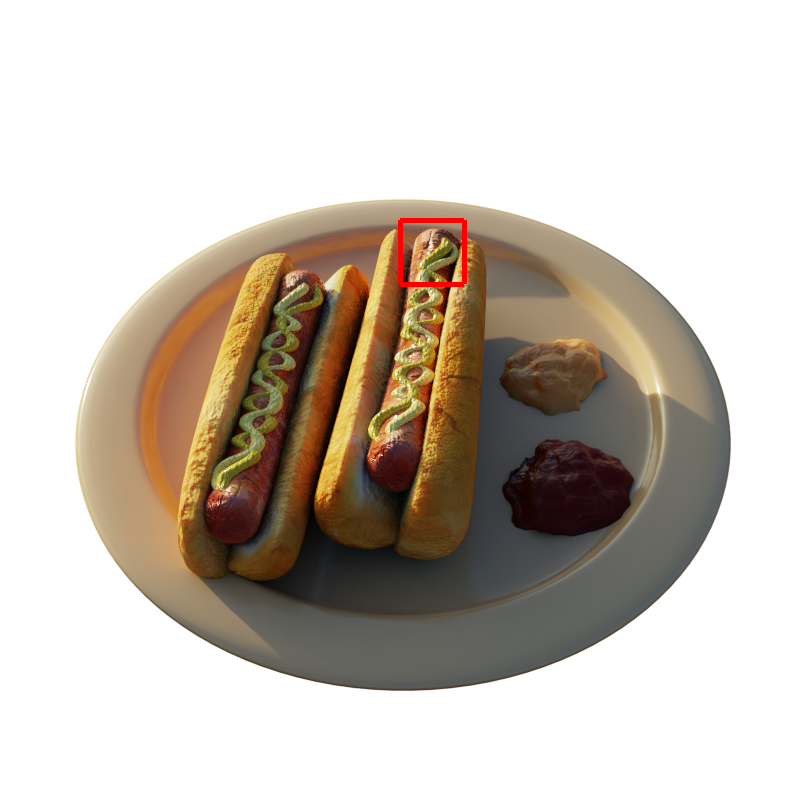} & 
    \textpyr{gt_hotdog_2_220_400_64.png}{}{}{}{} & 
    \textpyr{jaxnerf_multiblender_hotdog_2_220_400_64.png}{0.709}{0.870}{0.804}{0.680} & 
    \textpyr{jaxnerf_multiblender_extras_hotdog_2_220_400_64.png}{0.833}{0.930}{0.945}{0.817} & 
    \textpyr{prenerf_multiblender_hotdog_2_220_400_64.png}{ \color{wincolor} 0.898}{ \color{wincolor} 0.956}{ \color{wincolor} 0.966}{ \color{wincolor} 0.965} \\ 
    \flatim{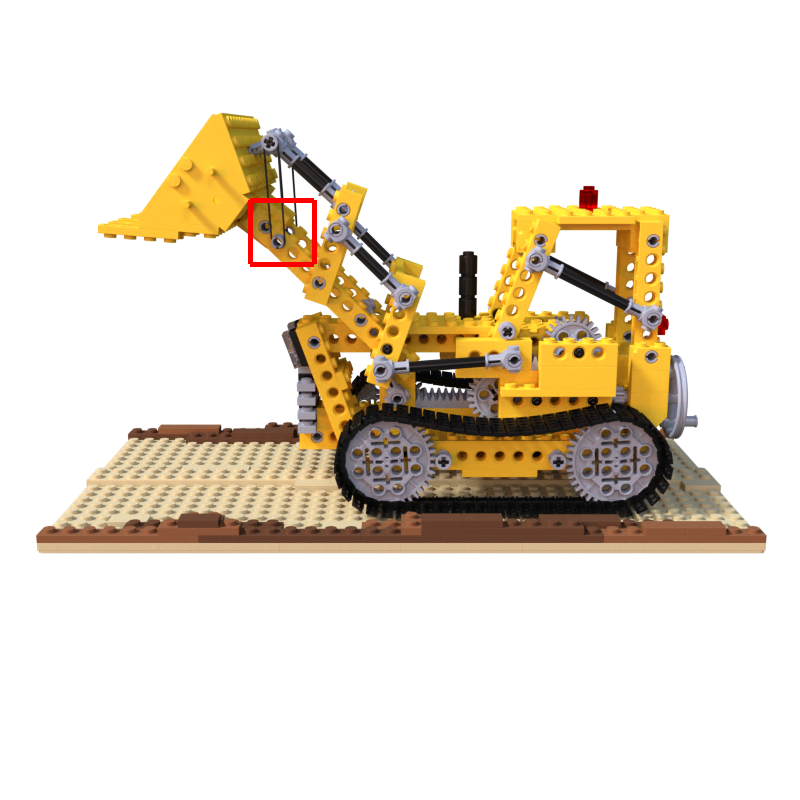} & 
    \textpyr{gt_lego_75_200_250_64.png}{}{}{}{} & 
    \textpyr{jaxnerf_multiblender_lego_75_200_250_64.png}{0.561}{0.621}{0.614}{0.682} & 
    \textpyr{jaxnerf_multiblender_extras_lego_75_200_250_64.png}{0.696}{0.742}{0.765}{0.882} & 
    \textpyr{prenerf_multiblender_lego_75_200_250_64.png}{ \color{wincolor} 0.830}{ \color{wincolor} 0.847}{ \color{wincolor} 0.795}{ \color{wincolor} 0.939} \\ 
    \flatim{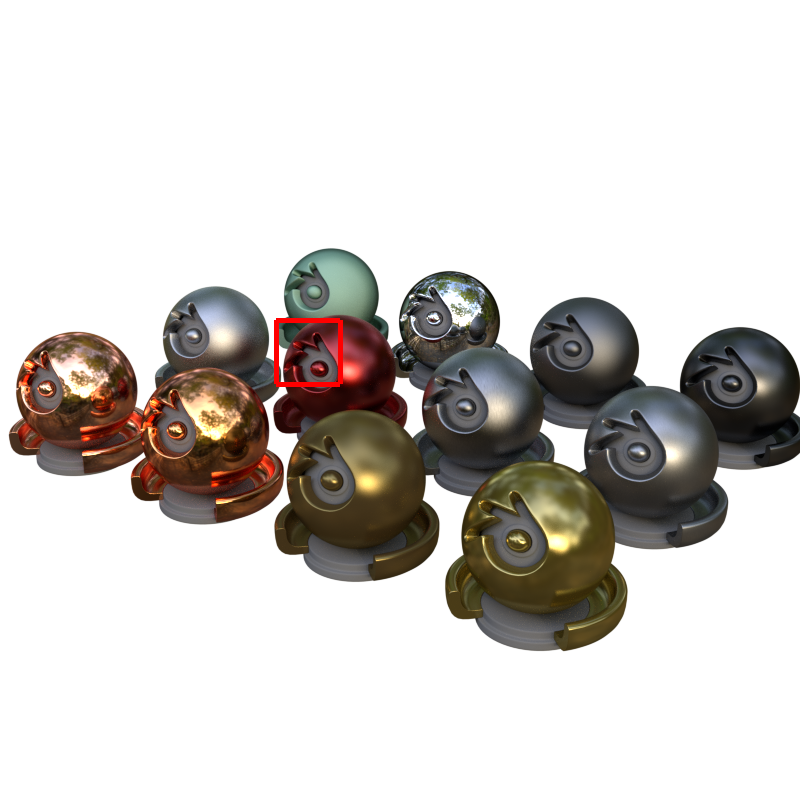} & 
    \textpyr{gt_materials_60_320_276_64.png}{}{}{}{} & 
    \textpyr{jaxnerf_multiblender_materials_60_320_276_64.png}{0.633}{0.763}{0.819}{0.804} & 
    \textpyr{jaxnerf_multiblender_extras_materials_60_320_276_64.png}{0.752}{0.838}{0.900}{0.935} & 
    \textpyr{prenerf_multiblender_materials_60_320_276_64.png}{ \color{wincolor} 0.821}{ \color{wincolor} 0.863}{ \color{wincolor} 0.902}{ \color{wincolor} 0.957} \\ 
    \flatim{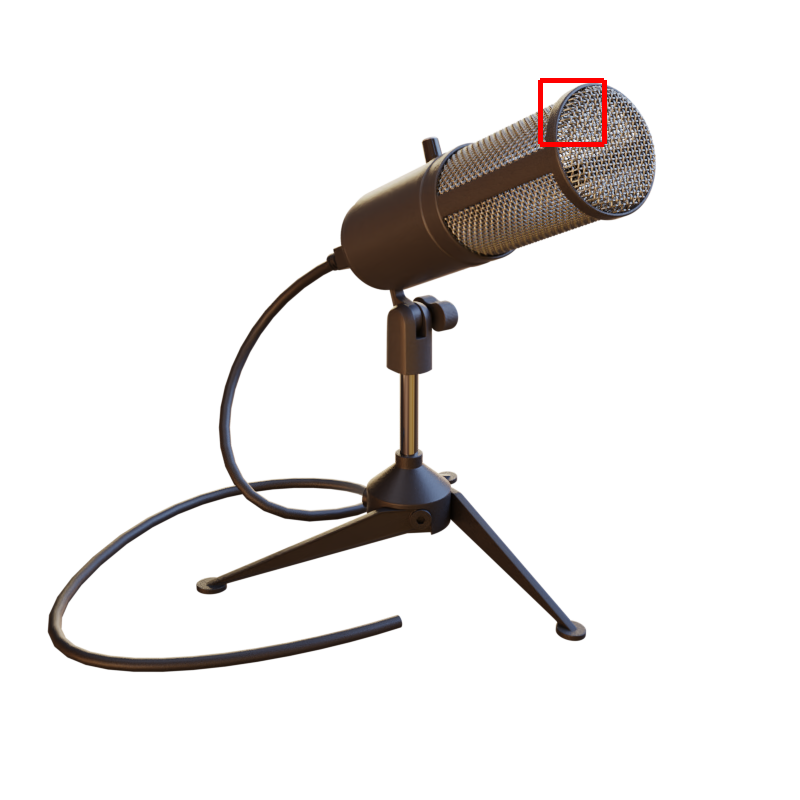} & 
    \textpyr{gt_mic_140_80_540_64.png}{}{}{}{} & 
    \textpyr{jaxnerf_multiblender_mic_140_80_540_64.png}{0.425}{0.600}{0.739}{0.838} & 
    \textpyr{jaxnerf_multiblender_extras_mic_140_80_540_64.png}{0.566}{0.745}{0.776}{0.924} & 
    \textpyr{prenerf_multiblender_mic_140_80_540_64.png}{ \color{wincolor} 0.793}{ \color{wincolor} 0.844}{ \color{wincolor} 0.846}{ \color{wincolor} 0.987} \\ 
    \flatim{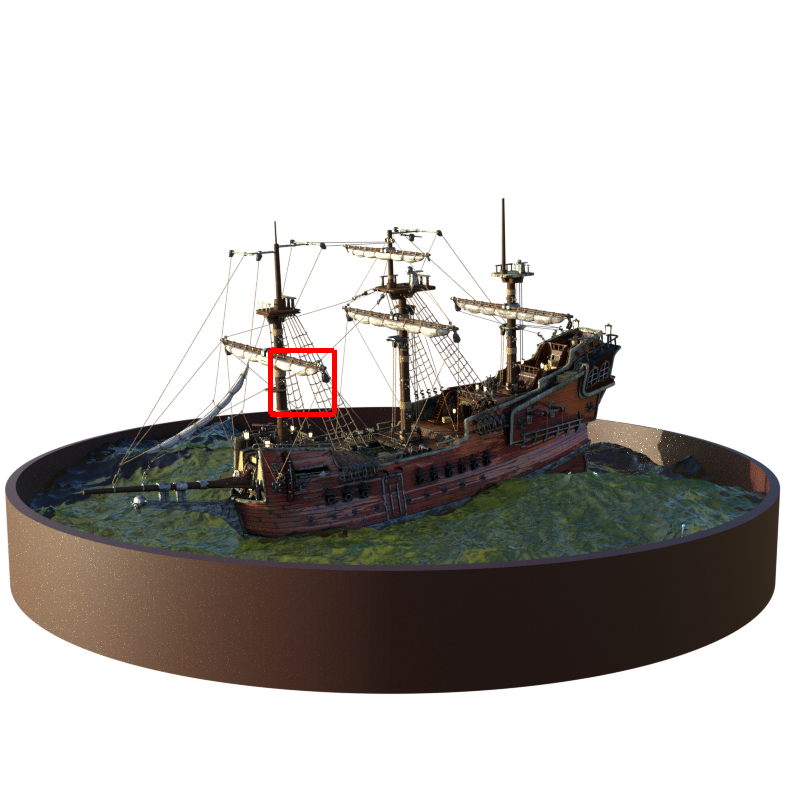} & 
    \textpyr{gt_ship_90_350_270_64.png}{}{}{}{} & 
    \textpyr{jaxnerf_multiblender_ship_90_350_270_64.png}{0.420}{0.618}{0.723}{0.755} & 
    \textpyr{jaxnerf_multiblender_extras_ship_90_350_270_64.png}{0.584}{0.810}{0.937}{0.907} & 
    \textpyr{prenerf_multiblender_ship_90_350_270_64.png}{ \color{wincolor} 0.823}{ \color{wincolor} 0.929}{ \color{wincolor} 0.963}{ \color{wincolor} 0.968} \\ 
    \multicolumn{2}{c|}{\footnotesize{Ground-Truth}} &
    \footnotesize{NeRF} & 
    \footnotesize{NeRF + Area, Center, Misc} & 
    \footnotesize{\shortmodelname}
    \end{tabular}
    \caption{
    Visualizations of the output renderings from \shortmodelnamelower compared to the ground truth, NeRF, and our improved version of NeRF, on test set images from the 8 scenes in our multiscale Blender dataset.
    We visualize a cropped region of each scene for better visualization, and render out that scene at 4 different resolutions, displayed as an image pyramid. The SSIM for each scale of each image pyramid truth is shown to its lower right, with the highest SSIM for each algorithm at each scale highlighted in red.
    }
    \label{fig:multi_vis}
\end{figure*}

 \begin{table*}[]
    \centering
    \small
    \begin{tabular}{l|cccccccc}
 & \multicolumn{8}{c}{PSNR} \\
 & \scenename{chair}  & \scenename{drums}  & \scenename{ficus}  & \scenename{hotdog}  & \scenename{lego}  & \scenename{materials}  & \scenename{mic}  & \scenename{ship} \\ \hline 
SRN~\cite{srn}                                                    &                    26.96  &                    17.18  &                    20.73  &                    26.81  &                    20.85  &                    18.09  &                    26.85  &                    20.60  \\
Neural Volumes~\cite{neuralvolumes}                               &                    28.33  &                    22.58  &                    24.79  &                    30.71  &                    26.08  &                    24.22  &                    27.78  &                    23.93  \\
LLFF~\cite{mildenhall2019}                                        &                    28.72  &                    21.13  &                    21.79  &                    31.41  &                    24.54  &                    20.72  &                    27.48  &                    23.22  \\
NSVF~\cite{liu2020neural} & 33.19 & 25.18 & 31.23 & 37.14 & 32.29 &  \cellcolor{red} 32.68 & 34.27 & 27.93 \\
NeRF (TF Implementation)~\cite{mildenhall2020}                    &                    33.00  &                    25.01  &                    30.13  &                    36.18  &                    32.54  &                    29.62  &                    32.91  &                    28.65  \\
NeRF (Jax Implementation)~\cite{jaxnerf2020github,mildenhall2020} &                    34.17  &                    25.08  &                    30.39  &                    36.82  &                    33.31  &                    30.03  &                    34.78  &                    29.30  \\
NeRF + Centered Pixels                                            &                    34.88  &                    25.17  &                    31.02  &                    37.13  &                    34.39  &                    30.50  &                    35.38  &                    29.95  \\
NeRF + Center, Misc.                                              &                    34.94  &                    25.19  &                    31.05  &                    37.15  &                    34.12  &                    30.47  &                    35.33  &                    29.95  \\
\hline
\shortmodelname                                                   & \cellcolor{orange} 35.14  & \cellcolor{red}    25.48  & \cellcolor{red}    33.29  & \cellcolor{orange} 37.48  & \cellcolor{red}    35.70  & \cellcolor{orange}    30.71  & \cellcolor{red}    36.51  & \cellcolor{yellow} 30.41  \\
\shortmodelname w/o Single MLP                                    &                    35.07  & \cellcolor{yellow} 25.28  & \cellcolor{yellow} 32.52  & \cellcolor{yellow} 37.34  & \cellcolor{yellow} 34.93  &                    30.38  &                    35.59  & \cellcolor{red}    30.55  \\
\shortmodelname w/o Misc.                                         & \cellcolor{red}    35.16  & \cellcolor{orange} 25.46  & \cellcolor{orange} 32.96  & \cellcolor{red}    37.55  & \cellcolor{orange} 35.68  & \cellcolor{yellow} 30.69  & \cellcolor{orange} 36.32  & \cellcolor{orange} 30.47  \\
\shortmodelname w/o \shortfeaturename                             & \cellcolor{yellow} 35.10  &                    25.23  &                    31.30  &                    37.17  &                    34.89  &  30.56  & \cellcolor{yellow} 35.75  &                    29.85  \\
\shortmodelname, Stopped Early &                    34.21  &                    25.23  &                    30.79  &                    36.89  &                    33.72  &                    29.86  &                    35.02  &                    29.44  \\

\multicolumn{9}{c}{} \\
 & \multicolumn{8}{c}{SSIM} \\
 & \scenename{chair}  & \scenename{drums}  & \scenename{ficus}  & \scenename{hotdog}  & \scenename{lego}  & \scenename{materials}  & \scenename{mic}  & \scenename{ship} \\ \hline 
SRN~\cite{srn}                                                    &                    0.910  &                    0.766  &                    0.849  &                    0.923  &                    0.809  &                    0.808  &                    0.947  &                    0.757  \\
Neural Volumes~\cite{neuralvolumes}                               &                    0.916  &                    0.873  &                    0.910  &                    0.944  &                    0.880  &                    0.888  &                    0.946  &                    0.784  \\
LLFF~\cite{mildenhall2019}                                        &                    0.948  &                    0.890  &                    0.896  &                    0.965  &                    0.911  &                    0.890  &                    0.964  &                    0.823  \\
NSVF~\cite{liu2020neural} & 0.968 & \cellcolor{orange} 0.931 &  0.973 & \cellcolor{yellow} 0.980 & 0.960 &  \cellcolor{red} 0.973 & 0.987 & 0.854 \\
NeRF (TF Implementation)~\cite{mildenhall2020}                    &                    0.967  &                    0.925  &                    0.964  &                    0.974  &                    0.961  &                    0.949  &                    0.980  &                    0.856  \\
NeRF (Jax Implementation)~\cite{jaxnerf2020github,mildenhall2020} &                    0.975  &                    0.925  &                    0.967  &                    0.979  &                    0.968  &                    0.953  &                    0.987  &                    0.869  \\
NeRF + Centered Pixels                                            & \cellcolor{yellow} 0.979  & 0.928  &                    0.971  & \cellcolor{yellow} 0.980  &                    0.973  &  0.956  & \cellcolor{yellow} 0.989  &                    0.877  \\
NeRF + Center, Misc.                                              & \cellcolor{yellow} 0.979  &                    0.927  &                    0.971  & \cellcolor{yellow} 0.980  &                    0.972  &  0.956  & \cellcolor{yellow} 0.989  &                    0.877  \\
\hline
\shortmodelname                                                   & \cellcolor{red}    0.981  & \cellcolor{red}    0.932  & \cellcolor{red}    0.980  & \cellcolor{red}    0.982  & \cellcolor{red}    0.978  & \cellcolor{orange}    0.959  & \cellcolor{red}    0.991  & \cellcolor{orange} 0.882  \\
\shortmodelname w/o Single MLP                                    & \cellcolor{orange} 0.980  & \cellcolor{yellow} 0.929  & \cellcolor{yellow} 0.977  & \cellcolor{orange} 0.981  & \cellcolor{orange} 0.976  &  0.956  & \cellcolor{yellow} 0.989  & \cellcolor{red}    0.883  \\
\shortmodelname w/o Misc.                                         & \cellcolor{red}    0.981  & \cellcolor{red}    0.932  & \cellcolor{orange} 0.979  & \cellcolor{red}    0.982  & \cellcolor{red}    0.978  & \cellcolor{orange}    0.959  & \cellcolor{red}    0.991  & \cellcolor{red}    0.883  \\
\shortmodelname w/o \shortfeaturename                             & \cellcolor{red}    0.981  & \cellcolor{yellow} 0.929  &                    0.972  & \cellcolor{orange} 0.981  & \cellcolor{yellow} 0.975  & \cellcolor{yellow} 0.958  & \cellcolor{orange} 0.990  & \cellcolor{yellow} 0.878  \\
\shortmodelname, Stopped Early &                    0.976  &                    0.927  &                    0.969  &                    0.979  &                    0.969  &                    0.954  &                    0.988  &                    0.869  \\

\multicolumn{9}{c}{} \\
 & \multicolumn{8}{c}{LPIPS} \\
 & \scenename{chair}  & \scenename{drums}  & \scenename{ficus}  & \scenename{hotdog}  & \scenename{lego}  & \scenename{materials}  & \scenename{mic}  & \scenename{ship} \\ \hline 
SRN~\cite{srn}                                                    &                    0.106  &                    0.267  &                    0.149  &                    0.100  &                    0.200  &                    0.174  &                    0.063  &                    0.299  \\
Neural Volumes~\cite{neuralvolumes}                               &                    0.109  &                    0.214  &                    0.162  &                    0.109  &                    0.175  &                    0.130  &                    0.107  &                    0.276  \\
LLFF~\cite{mildenhall2019}                                        &                    0.064  &                    0.126  &                    0.130  &                    0.061  &                    0.110  &                    0.117  &                    0.084  &                    0.218  \\
NSVF~\cite{liu2020neural} & 0.043 & 0.069 &  \cellcolor{red} 0.017 & \cellcolor{red} 0.025 &  0.029 &  \cellcolor{red} 0.021 & \cellcolor{orange} 0.010 &  0.162 \\
NeRF (TF Implementation)~\cite{mildenhall2020}                    &                    0.046  &                    0.091  &                    0.044  &                    0.121  &                    0.050  &                    0.063  &                    0.028  &                    0.206  \\
NeRF (Jax Implementation)~\cite{jaxnerf2020github,mildenhall2020} &                    0.026  &                    0.071  &                    0.032  &                    0.030  &                    0.031  &                    0.047  &                    0.012  &                    0.150  \\
NeRF + Centered Pixels                                            & \cellcolor{yellow} 0.022  &                    0.069  &                    0.028  & 0.028  & \cellcolor{yellow} 0.026  & 0.043  & \cellcolor{orange} 0.010  &                    0.143  \\
NeRF + Center, Misc.                                              & \cellcolor{yellow} 0.022  &                    0.069  &                    0.028  & 0.028  &                    0.027  &                    0.044  & \cellcolor{yellow} 0.011  &                    0.142  \\
\hline
\shortmodelname                                                   & \cellcolor{orange} 0.021  & \cellcolor{red}    0.065  & \cellcolor{orange}    0.020  & \cellcolor{yellow} 0.027  & \cellcolor{red}    0.021  & \cellcolor{orange}    0.040  & \cellcolor{red}    0.009  & \cellcolor{yellow} 0.138  \\
\shortmodelname w/o Single MLP                                    & \cellcolor{yellow} 0.022  & \cellcolor{yellow} 0.067  &  0.023  &  0.028  & \cellcolor{orange} 0.024  &                    0.044  & \cellcolor{yellow} 0.011  & \cellcolor{red}    0.135  \\
\shortmodelname w/o Misc.                                         & \cellcolor{orange} 0.021  & \cellcolor{orange} 0.066  & \cellcolor{yellow} 0.022  & \cellcolor{orange}    0.026  & \cellcolor{red}    0.021  & \cellcolor{orange}    0.040  & \cellcolor{red}    0.009  & \cellcolor{orange} 0.136  \\
\shortmodelname w/o \shortfeaturename                             & \cellcolor{red}    0.020  &                    0.068  &                    0.027  &  0.028  & \cellcolor{orange} 0.024  & \cellcolor{yellow} 0.041  & \cellcolor{red}    0.009  &                    0.142  \\
\shortmodelname, Stopped Early &                    0.027  &                    0.074  &                    0.035  &                    0.031  &                    0.035  &                    0.046  &                    0.013  &                    0.155  \\

    \end{tabular}
    \caption{Per-scene results on the test set images of the (single-scale) Blender dataset from Mildenhall~\etal~\cite{mildenhall2020}}
    \label{tab:blender_results}
\end{table*}

\end{document}